\documentclass{article}
\usepackage{jfrExamplee}
\usepackage{graphicx}
\usepackage{apalike}
\usepackage{setspace}

\usepackage{psfrag}
\usepackage{amsmath}
\usepackage{amssymb}
\usepackage{graphicx}
\usepackage{subfigure}
\usepackage{verbatim}
\usepackage[thinlines,thiklines]{easybmat}
\usepackage{latexsym}
\usepackage[dvipsnames]{xcolor}
\usepackage{cite}
\usepackage{stmaryrd}
\usepackage{multirow}
\usepackage{textcomp}
\usepackage[ruled]{algorithm2e}
\usepackage[T1]{fontenc}

\newbox\tempbox

\newcommand{\bs}[1]{\ensuremath{{\boldsymbol{#1}}}}

\def\atan2{\mathrm{atan2}}

\title{Development of An Autonomous Bridge Deck Inspection Robotic System}

\author{
Hung M.~La\thanks{ Hung La is the Director of Advanced Robotics and Automation (ARA) Lab (http://ara.cse.unr.edu/) and an Assistant Professor of Department of Computer Science and Engineering, University of Nevada, Reno. } \\
Department of Computer Science and Engineering\\
University of Nevada\\
Reno, NV 89557, USA. \\
\texttt{hla@unr.edu} \\
\And
Nenad Gucunski \\
Department of Civil and Environmental Engineering, Rutgers University \\
Piscataway, NJ 08854, USA. \\
\texttt{gucunski@rci.rutgers.edu} \\
\AND
 Kristin Dana \\
Department of Electrical and Computer Engineering, Rutgers University \\
Piscataway, NJ 08854, USA. \\
\texttt{kdana@ece.rutgers.edu} \\
\And
Seong-Hoon Kee \\
Department of Architectural Engineering, Dong-A University, \\
 Busan, S. Korea \\
\texttt{shkee@dau.ac.kr} \\
}

\begin{document}

\maketitle

\begin{abstract}
The threat to safety of aging bridges has been recognized as a critical concern to the general public due to the poor condition of many bridges in the U.S.  Currently, the bridge inspection is conducted manually, and it is not efficient  to identify bridge condition deterioration in order to facilitate implementation of appropriate maintenance or rehabilitation procedures. In this paper, we report a new development of the autonomous mobile robotic system for bridge deck inspection and evaluation. The robot is integrated with several nondestructive evaluation (NDE) sensors and a  navigation control algorithm to allow it to accurately and autonomously maneuver on the bridge deck to collect visual images and conduct NDE measurements. The developed robotic system can reduce the cost and time of the bridge deck data collection and inspection. For efficient bridge deck monitoring, the crack detection algorithm to build the deck crack map is presented in detail. The impact-echo (IE), ultrasonic surface waves (USW) and electrical resistivity (ER) data collected by the robot are analyzed to generate the delamination, concrete elastic modulus, corrosion maps of the bridge deck, respectively. The presented robotic system has been successfully deployed to inspect numerous bridges in more than ten different states in the U.S.
\end{abstract}


\section{Introduction}
\label{Intro}
\begin{figure}[htp!]
\centering
\includegraphics[width=10cm]{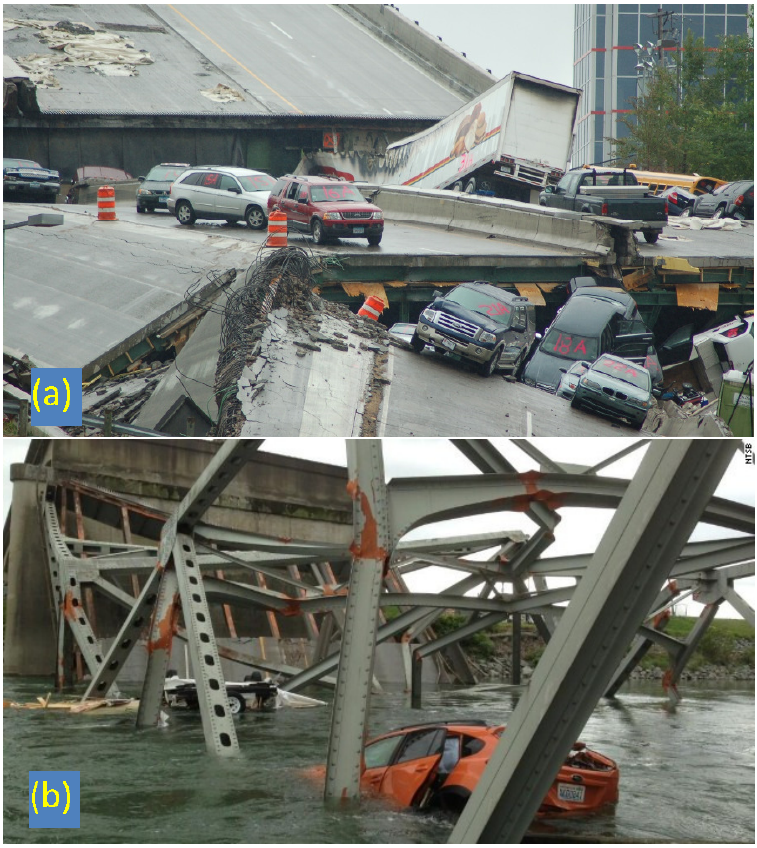}
\caption{Bridge collapse: (a) at Minnesota state in 2007, source: Wikipedia \cite{Bridge_collapse_I35W}; (b) at Washington state in 2013, source: Urbansplatter \cite{Bridge_collapse_Washington}.}
\label{Bridge Collapse}
\end{figure}

One of the biggest challenges the United States faces today is infrastructure like-bridges inspection and maintenance. The threat to safety of aging bridges has been recognized as a growing problem of national concern to the general public. There are currently more than 600,000 bridges in the U.S., the condition of which is critical for the safety of the traveling public and economic vitality of the country. According to the National Bridge Inventory there are about 150,000 bridges through the U.S. that are structurally deficient or functionally obsolete due to various mechanical and weather conditions, inadequate
maintenance, and deficiencies in inspection and evaluation \cite{FHWA2008}, and this number is growing. Numerous bridges collapsed recently (Fig. \ref{Bridge Collapse}) have raised a strong call for efficient bridge inspection and evaluation. The cost of maintenance and rehabilitation of the deteriorating bridges is immense. The cost of repairing and replacing deteriorating highway bridges in U.S. was estimated to be more than \$140 billions in 2008~\cite{ReportCard2009}. Condition monitoring and timely
implementation of maintenance and rehabilitation procedures are needed to reduce future costs associated with bridge management. Application of nondestructive evaluation (NDE) technologies is one of the effective ways to monitor and predict bridge deterioration.  A number of NDE technologies are currently used in bridge deck evaluation, including impact-echo (IE), ground penetrating radar (GPR), electrical resistivity (ER), ultrasonic surface waves (USW) testing, visual inspection, etc.~\cite{Gucunski2010TRB}. 

Automated bridge inspection is the safe, efficient and reliable way to evaluate condition of the bridge \cite{Mohammad_SIE2009}. Thus, robotics and automation technologies have increasingly gained attention for bridge inspection, maintenance, and rehabilitation. Mobile robot- or vehicle-based inspection and maintenance systems are developed for vision based crack detection and maintenance of highways and tunnels ~\cite{Velinsky1993,Lorenc2000,Seung07}. A robotic system for underwater inspection of bridge piers is reported in \cite{DeVault2000}. An adaptive control algorithm for a bridge-climbing robot is developed \cite{Liu2013}. Additionally, robotic systems for steel structured bridges are developed \cite{Wang2007, Mazumdar2009, Cho2013}. In one case, a mobile manipulator is used for bridge crack inspection ~\cite{Pi02}. A bridge inspection system that includes a specially designed car with a robotic mechanism and a control system for automatic crack detection is reported in ~\cite{Jeong08,Lee2008,Je09}. Similar systems are reported in~\cite{Lim2011, Lim_TASE_2014} for vision-based automatic crack detection and mapping and ~\cite{Yu2007, Prasanna2014, Ying2010, Zou2012, Fujita2011, Nishi2012, Jahan2013b} to detect cracks on a tunnel. Robotic rehabilitation systems for concrete repair and automatically filling the delamination inside bridge decks have also been reported in ~\cite{Chamber2002,LiuCASE2013}. 
\begin{figure*}[t!]
	\centering
	\includegraphics[width=\textwidth]{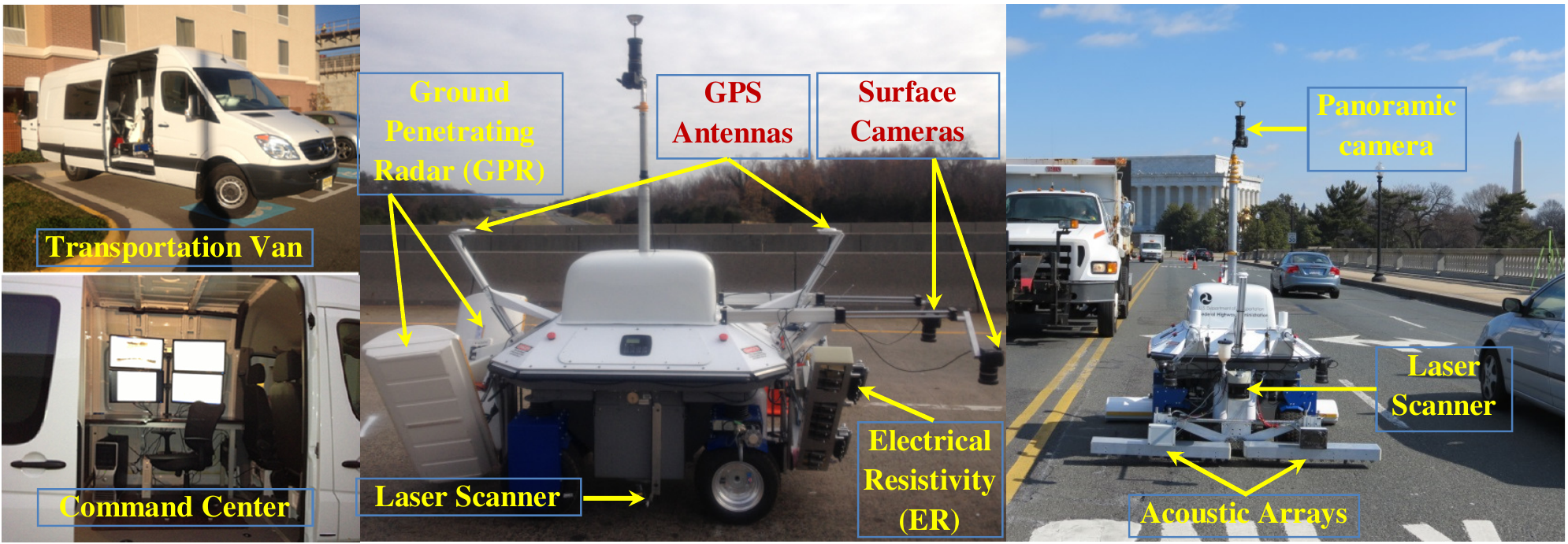}
	\caption{The robotic NDE bridge deck
inspection system: (Left) Transportation van and the command center of the robot; (Middle) The side view of the robot in the idle mode; (Right) The front view of the robot in the working mode  on the Arlington Memorial bridge, Washington DC, USA in 2014.}
	\label{robot}
\end{figure*}

Most of abovementioned works classify, measure, and detect cracks. However, none of these works studies the global mapping of cracks, delamination,  elastic modulus and corrosion of the bridge decks based on different NDE technologies. Difference to all of the above mentioned works, in this paper we focus on the development of a real world practical robotic system which integrates advanced NDE technologies for the bridge inspection. The robot can autonomously and accurately maneuver on the bridge to collect NDE data including high resolution images, impact-echo (IE), ultrasonic surface waves (USW), electrical resistivity (ER) and ground penetrating radar (GPR). The data is stored in the onboard robot computers as well as wirelessly transferred to the command center on the van for online data processing. 

Compared to the current manual data collection technologies, the developed robotic system (Fig. \ref{robot}) can reduce the cost and time of the bridge deck data collection. More importance, there is no safety risks since the robot can autonomously travel and collect data on the bridge deck without human operators. Moreover, advanced data analysis algorithms are proposed by taking into account the advantages of the accurate robotic localization and navigation to provide the high resolution bridge deck image, crack map, and delamination, elastic modulus and corrosion maps of the bridge deck, respectively. This creates the ease of bridge condition assessments and monitor in timely manner. The initial report of the proposed robotic system was published in \cite{La_IROS_2014}.

\begin{figure*}[htb!]
	\centering
	\includegraphics[width=\textwidth]{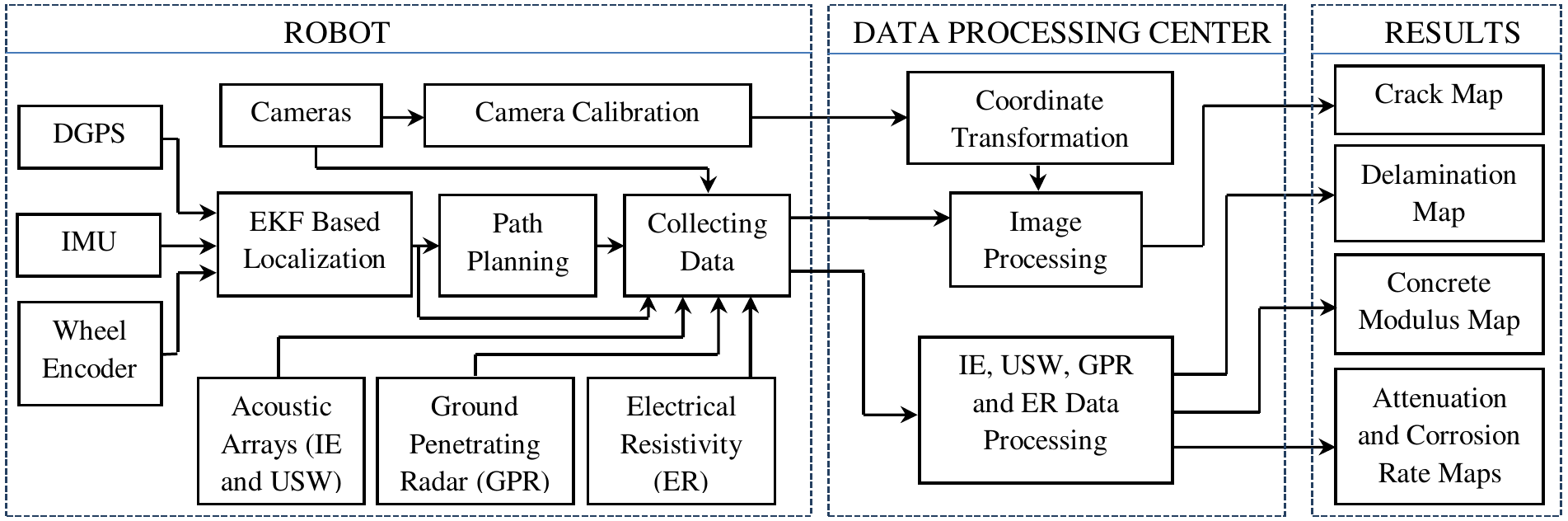}
	\caption{The working principle of the robotic bridge deck
inspection system.}
	\label{robot_diagram}
\end{figure*}

The rest of the paper is organized as follows. In the next section, we describe the overall design of the robotic system, the software integration of NDE sensors, and autonomous navigation design. In Section \ref{CDA}, we  present robotic data collection and analysis. The testing results and field deployments are presented in Section~\ref{Results}. Finally, we provide conclusions from the current work and discuss the future work in Section~\ref{Con}.
 
\section{The Robotic System for Bridge Deck Inspection and Evaluation} 
\label{ORS}
\subsection{Overview of the Robotic System}
Fig.~\ref{robot} shows the autonomous robotic NDE system. The mobile platform is a Seekur robot from Adept Mobile Robot Inc. The Seekur robot is an electrical all-wheel driving and steering platform that can achieve highly agile maneuvers on bridge decks. The mobile robot has been significantly modified and equipped with various sensors, actuators, and computing devices. The localization and motion planning sensors include two RTK GPS units (from Novatel Inc.), one front- and two side-mounted laser scanners (from Sick AG and Hokuyo Automation Co., respectively), and one IMU sensor (from Microstrain Inc.) The onboard NDE sensors
include two ground penetration radar (GPR) arrays, two seismic/acoustic sensor arrays, four electrical resistivity (Wenner) probes, two high-resolution
surface imaging cameras and a 360-degree panoramic camera. The details of the system mechatronic design and the autonomous robotic localization algorithm based Extended Kalman Filter (EKF) are provided in~\cite{LaCASE2013,LaTmech2013}.

Three embedded computers  are integrated inside the robot. One computer runs Robotic Operating System (ROS) nodes for the
robot localization, navigation and motion planning tasks. The other two computers are used for the NDE sensors integration and data collection. High-speed Ethernet connections are used among these computers. Each computer can also be reached individually through a high-speed wireless communication from remote computers. The NDE
data are transmitted in real-time to the remote computers in the command center for visualization and data analysis purposes (see Fig. \ref{robot_diagram}). 

\subsection{Autonomous Navigation Design for the Robotic System}
\begin{figure}[htb!]
	\centering
	\includegraphics[width=10cm]{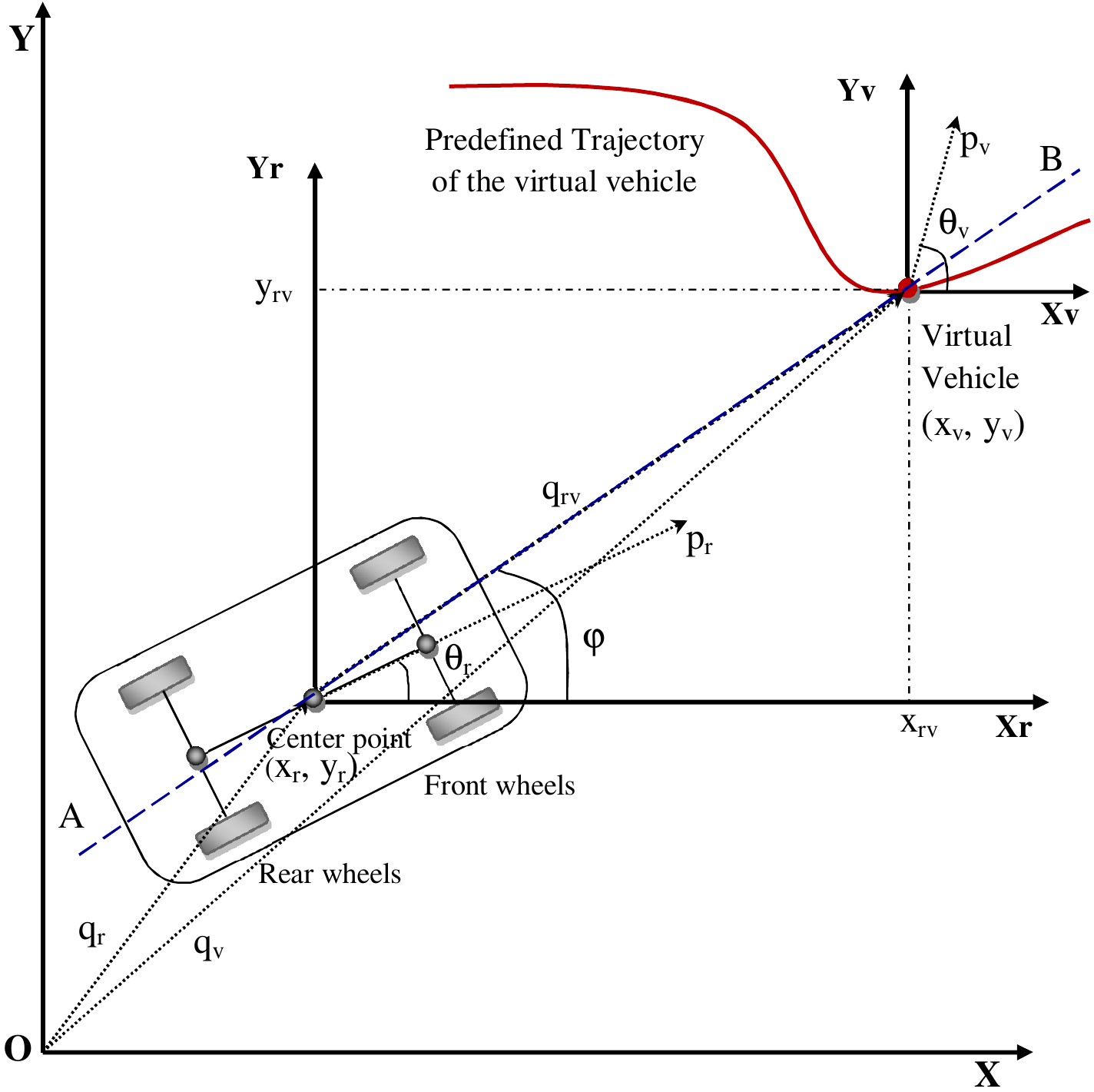}
	\caption{A mobile robot tracks a moving target/virtual robot.}
	\label{tracking}
\end{figure}

The robot navigation is developed based on artificial potential field approach. We design a virtual robot to move in the predefined trajectory so that it can cover the entire desired inspection area. This virtual robot generates an artificial attractive force to attract the robot to follow as illustrated in Fig.  $\ref{tracking}$ with notations defined as follows.

$q_{r} = [x_r \ y_r]^{T} \in R^{2}, p_{r} = \dot{q}_r \in R^{2},\theta_{r} \in R^{1} $ are position, velocity, and heading of the mobile robot at time $t$,  respectively. Note that the robot's pose ($q_{r}, \theta_r$) is obtained by the localization algorithm using an Extended Kalman Filter (EKF)   \cite{LaCASE2013,LaTmech2013}. $q_{v} = [x_{v} \ y_{v}]^{T} \in R^{2}, p_{v} \in R^{2},\theta_{v} \in R^{1} $ are position, velocity, and heading of the virtual robot at time $t$, respectively. $\varphi$ is the relative angle between the mobile robot and the virtual one, and it is computed as $\varphi = atan2(y_{rv},x_{rv})$.

Let $q_{rv} = [x_{rv}, \ y_{rv}]^{T}$ be the relative position between a mobile robot and  a virtual robot with $	x_{rv} = x_{v}- x_{r}$ and $y_{rv} = y_{v}- y_{r}$.  The control objective is to regulate $\|q_{rv}\|$ to  zero as soon as possible. This means that $q_{r} = q_{v}$ and $p_{r} = p_{v}$. To achieve such a controller design, the potential field approach  \cite{Ge2000, Huang2008} is utilized to design an attractive potential function as follows 
\begin{equation}
		V_{a} = \frac{1}{2}\lambda \|q_{rv}\|^{2} = \frac{1}{2}\lambda q^{T}_{rv}q_{rv},
		\label{potential}
\end{equation}
here $\lambda$ is a small positive constant for the attractive  potential field function, and in our experiment we select $\lambda = 0.05$.

To track the virtual robot, we design the velocity controller of the mobile robot as
\begin{equation}
p^{d}_{r}= \nabla_{q_{rv}}V_a = \lambda q_{rv},
\label{stationary pal}
\end{equation}
where operator $\nabla_{q_{rv}}V_a$ represents the gradient calculation of scalar $V_a$ along vector $q_{rv}$.  The velocity controller given by~($\ref{stationary pal}$) is with respect to the stationary target. For tracking a moving target (e.g., the virtual robot), we can obtain the desired velocity $p^{d}_{r}$ of the mobile robot as
\begin{equation}
		p^{d}_{r} =  p_{v} + \lambda q_{rv}.
		\label{moving pal}
\end{equation}

We have the following theorem to show that the designed velocity controller given by~($\ref{moving pal}$) will allow the mobile robot to follow the virtual one.

\textbf{Theorem 1.}
The designed velocity controller given by~($\ref{moving pal}$) allows  the  mobile robot ($q_{r}, p_{r}$) to follow a virtual robot ($q_{v}, p_{v}$).

\textit{The proof of Theorem 1 is given in the Appendix.}

Now we can  extend the result given by~($\ref{moving pal}$) to the holonomic robot control through the design of linear velocity and heading controllers ($v^{d}_{r}, \theta^{d}_{r}$).  Taking $norm$ both sides of  Equ. ($\ref{moving pal}$) with the fact that $\|\vec{x}\vec{y}\| = xycos(\theta_{xy})$, here $\theta_{xy}$ is angle of two vectors $\vec{x}$ and $\vec{y}$, we obtain
\begin{eqnarray}
		v^{d}_{r} = \|p^{d}_{r}\| & = & 
  	\|	p_{v} + \lambda q_{rv}\|
  	\nonumber \\
		& = &
    \sqrt{\|p_{v}\|^{2} + 2\lambda \|q_{rv}\|\|p_{v}\|
		cos(\theta_{v} - \varphi) + \lambda^{2} \|q_{rv}\|^{2}}.
		\nonumber
\end{eqnarray}

\begin{figure*}[t!]
\centering
\includegraphics[width=\textwidth]{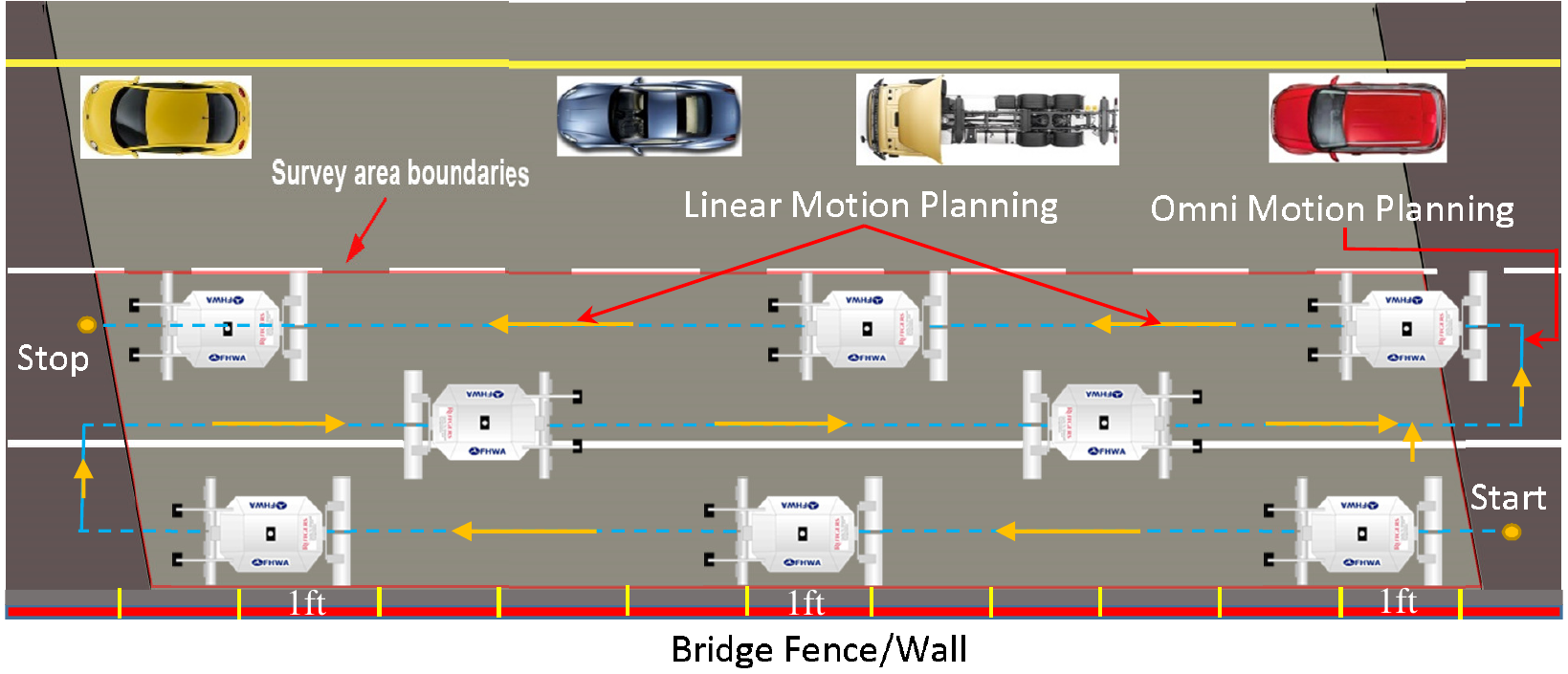}
\caption{Robot path planning on a bridge where the Start and Stop locations are preselected using GPS.}
\label{path planning}
\end{figure*} 

It is also desirable to have the equal projected velocities of the virtual and actual robot perpendicular to the line AB along their centers as shown in Fig.  \ref{tracking}. Therefore, we obtain the following relationship
\begin{equation}
		\|p_{r}\| sin(\theta_{r} - \varphi)=  \|p_{v}\| sin(\theta_{v} - \varphi).
		\label{equivelent pal}
\end{equation}
By dividing both sides of Equation ($\ref{equivelent pal}$) with $\|p_{r}\|$ and taking $arcsin$ we obtain the heading controller for the mobile robot as
\begin{equation}
		\theta^{d}_{r} =  \varphi + sin^{-1}(\frac{\|p_{v}\|sin(\theta_{v} - \varphi)}{\|p_{r}\|}).
		\label{theta al}
\end{equation}
However, we can see that $\frac{\|p_{v}\|sin(\theta_{v} - \varphi)}{\|p_{r}\|}$ could return bigger than 1, and it is not valid to compute $\theta^{d}_{r}$ given by Equ. (\ref{theta al}). Therefore we need to design $\|p_r\| \geq \|p_v\|$ so that  $\frac{\|p_{v}\|sin(\theta_{v} - \varphi)}{\|p_{r}\|} \leq 1$. One possible solution is taking the absolute value of the angle  $(\theta_{v} - \varphi)$, or this results
\begin{equation}
		v^{d}_{r} = \sqrt{\|p_{v}\|^{2} + 2\lambda_{1}\|q_{rv}\|\|p_{v}\| |cos(\theta_{v} - \varphi)| + \lambda^{2}_{1}\|q_{rv}\|^{2}}.
		\label{final pal update}
\end{equation}

We can summarize the linear motion navigation algorithm for the autonomous robot as:
\begin{equation}
			\left\{ \begin{array}{l}
										\varphi = atan2(y_{rv},x_{rv})\\
											v^{d}_{r} = \sqrt{\|p_{v}\|^{2} + 2\lambda \|q_{rv}\|\|p_{v}\| |cos(\theta_{v} - \varphi)| + \lambda^{2}\|q_{rv}\|^{2}}\\
									\theta^{d}_{r} =  \varphi + sin^{-1}(\frac{\|p_{v}\|sin(\theta_{v} - \varphi)}{\|p_{r}\|}).
							\end{array} 
			\right.
\label{nav_algorithm}
\end{equation}

As required the robot has to travel closely to the bridge curve/fence to collect entire bridge deck data (see Fig. \ref{path planning}), the designed distance between the robot side and the bridge curve is 1 foot ($\approx$ 30cm). Note that the size of the robot associated with NDE sensors is large (2m $\times$ 2m) which is similar to the size of the small sedan car,  therefore to avoid curve hitting when the robot turns to start another scan line (Fig. \ref{path planning}), the omni-motion navigation algorithm is developed to allow the robot to move to the predefined safe locations on the bridge while the robot still keeps its current heading/orientation. 

Let $(x_{i}{^s}, y_{i}{^s})$ be a safe $i$th location  on the bridge deck.  We have relative distance between the robot and the safe location along $x$ and $y$ as:
\begin{equation}
			\left\{ \begin{array}{l}
										\Delta x_{i}^{sr}(t) = x_{i}{^s} - x_r(t) \\
										\Delta y_{i}^{sr}(t) = y_{i}{^s} - y_r(t).
							\end{array} 
			\right.
\label{re_xy}
\end{equation}

Then, we have error motion controls along $x$ and $y$ as
\begin{equation}
\left[ \begin{array}{c} e_x(t) \\ e_y(t) \end{array} \right] = \underbrace{\begin{bmatrix} cos(\Phi) & sin(\Phi) \\ -sin(\Phi) & cos(\Phi) \end{bmatrix}}_{R} \times \left[ \begin{array}{c} \Delta x_{i}^{sr}(t) \\ \Delta y_{i}^{sr}(t) \end{array} \right],
\label{motion control matrix}
\end{equation} 
here $R$ is the rotation matrix, and $\Phi$ is the relative angle between the robot position and the safe location, and $e_x(t)$ and $e_y(t)$ are the error motion controls along $x$ and $y$, respectively. Now we can map our error motion controls to the speed control along $x$ and $y$ of the robot.
\begin{equation}
			\left\{ \begin{array}{l}
										V_{x_r}(t) = K^{P}_{x}e_x(t) + K^{D}_{x}[e_x(t) - e_x(t-1)] \\
										V_{y_r}(t) = K^{P}_{y}e_y(t)+ K^{D}_{y}[e_y(t) - e_y(t-1)].
							\end{array} 
			\right.
\label{map_xy}
\end{equation}
here, $(K^{P}_{x}, K^{D}_{x})$ and $(K^{P}_{y}, K^{D}_{y})$ are the gains of the PD controller along $x$ and $y$ respectively. 

We can design the gains of the PD controller as follows.

Let $V^{d}_{x_r}, V^{d}_{y_r}$ be the desired velocities of the robot along $x$ and $y$, respectively. We now can select the gains $(K^{P}_{x}, K^{P}_{y})$ as 
\begin{equation}
			\left\{ \begin{array}{l}
										K^{P}_{x} = \frac{V^{d}_{x_r}}{|x_{i}{^s} - x_r(0)|} \\
										K^{P}_{y} = \frac{V^{d}_{y_r}}{|y_{i}{^s} - y_r(0)|}
							\end{array} 
			\right.
\label{P gain selection}
\end{equation}
where $(x_r(0), y_r(0))$ is the initial position of the robot when it starts to move to the safe location. Note that $|x_{i}{^s} - x_r(0)| \neq 0$ and $|y_{i}{^s} - y_r(0)| \neq 0$. The gains $(K^{D}_{x}, K^{D}_{y})$ can be selected as		$K^{D}_{x} = K^{P}_{x} + K; K^{D}_{y} = K^{P}_{x} + K$, where $K$ is a constant with $0<K<1$ and in our experiment we select $K = 0.1$.

\section{Bridge Deck Data Collection and Analysis}
\label{CDA}

The robot autonomously navigates within the predefined survey area on the bridge deck as shown in Fig. \ref{path planning}. The linear motion planning algorithm (\ref{nav_algorithm}) allows the robot to move straightly along each scan on the deck. The robot can cover 6ft wide in each scan. Once the robot finish the current scan it will move to the other scan until the entire survey area is completely scanned (Fig. \ref{path planning}). At the end of each scan, the omni motion planning algorithm (\ref{map_xy}) is activated to enable the robot to move to the other scan while avoid hitting the bridge fence/wall (Fig. \ref{path planning}).

The robot can work in two different modes: non-stop mode, stop-move mode. In the non-stop mode, the robot can move continuously with the speed up to 2m/s, and only GPR and camera, which do not require physical touch on the bridge deck, are used to collect data. In the stop-move mode, the robot moves (with speed 0.5m/s) and stops at every certain distance (i.e., 2ft or ~60cm) to collect all four different NDE data (GPR, IE, USW, ER) and visual data. The reason for stopping is that the robot has to deploy the acoustic arrays and the ER probes to physically touch the bridge deck in order to collect data. 
As shown in Fig. \ref{path planning}, the Start and Stop positions are pre-selected by using high accuracy Differential GPS system (less than 2cm error). These Start and Stop positions are used to determine if the scan is completed.

\subsection{NDE Sensors}

\begin{figure}[htp!]
\centering
\includegraphics[width=\columnwidth]{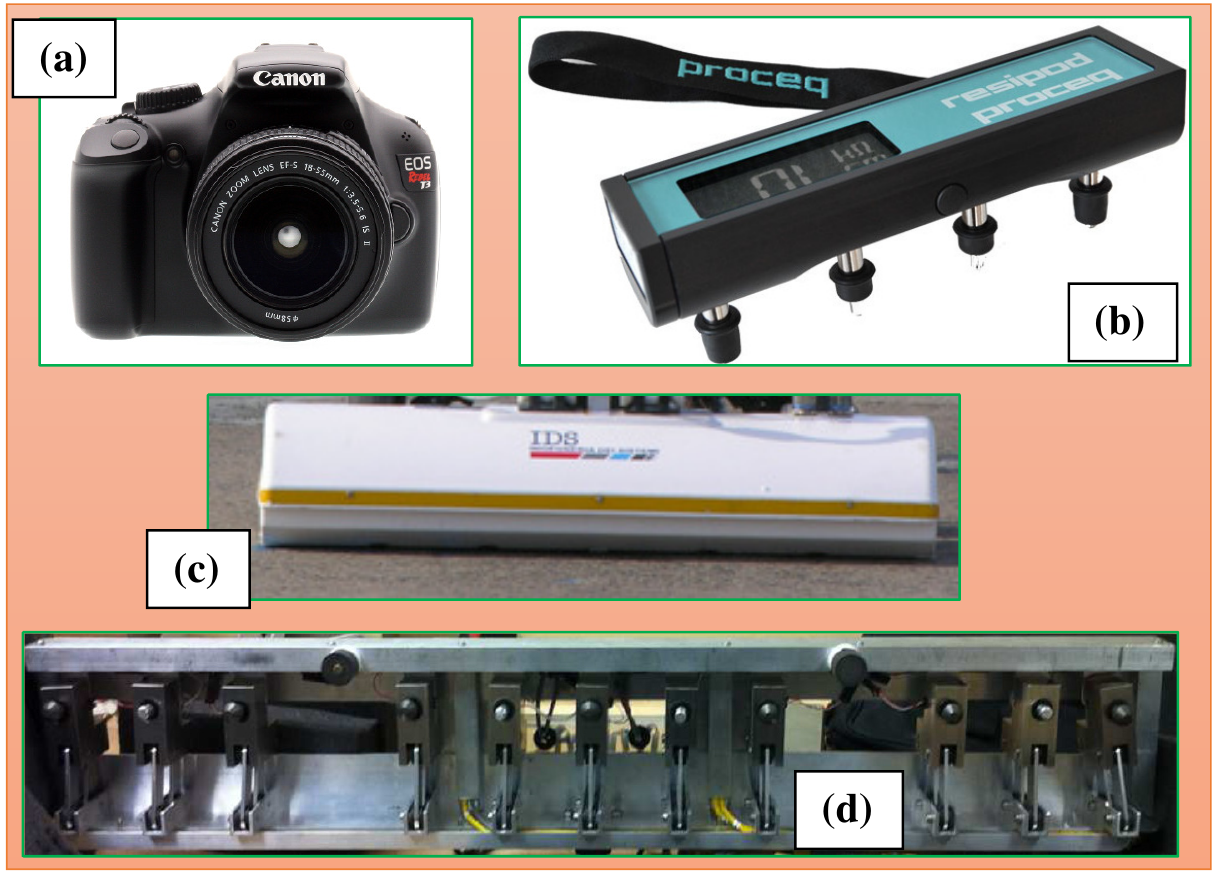}
\caption{NDE sensors: (a) Camera; (b) Electrical Resistivity (ER) Probe; (c) Ground Penetrating Radar (GPR); and (d) Acoustic array consists of Impact Echo (IE) and Ultrasonic Surface Wave (USW).}
\label{NDE sensors}
\end{figure}

There are four different NDE sensor technologies integrated with the robot including GPR, ER, IE and USW as shown in Fig.\ref{NDE sensors}.
GPR is a geophysical method that uses radar pulses to image the subsurface and describe the condition such as delamination. The IE method is a seismic resonant method that is primarily used to detect and characterize delamination (hidden cracks or vertical cracks) in bridge decks with respect to the delamination depth, spread and severity. USW technique is an offshoot of the spectral analysis of surface waves (SASW) method used to evaluate material properties (elastic moduli) in the near-deck surface zone. ER sensor measures concrete's electrical resistivity, which reflects the corrosive environment of bridge decks. The detail of these NDE technologies were presented in the previous work \cite{La_ISARC_2014, La_Springer_2015}.

\subsection{Visual Data}

\begin{figure}[htp!]
\centering
\includegraphics[width=\columnwidth]{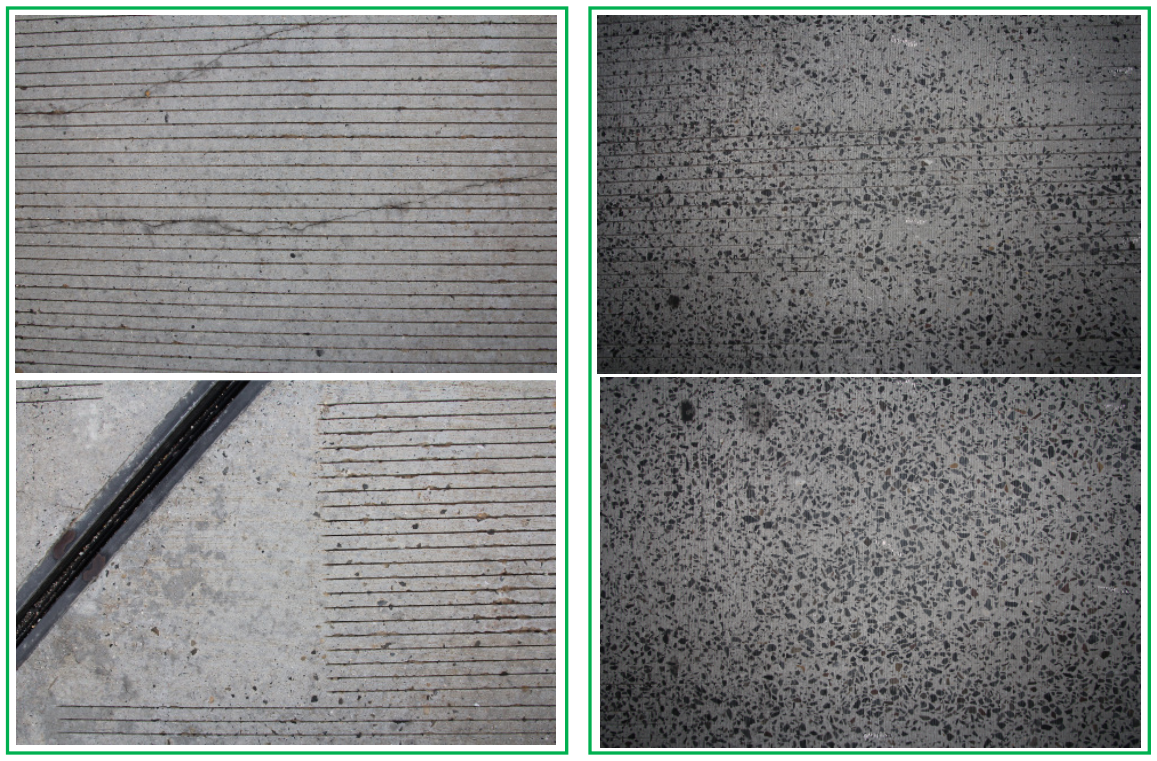}
\caption{ Sample images collected by the robot: (a) at day time. (b) at night time.}
\label{image samples}
\end{figure} 

\begin{figure}[htp!]
\centering
\includegraphics[width=\columnwidth]{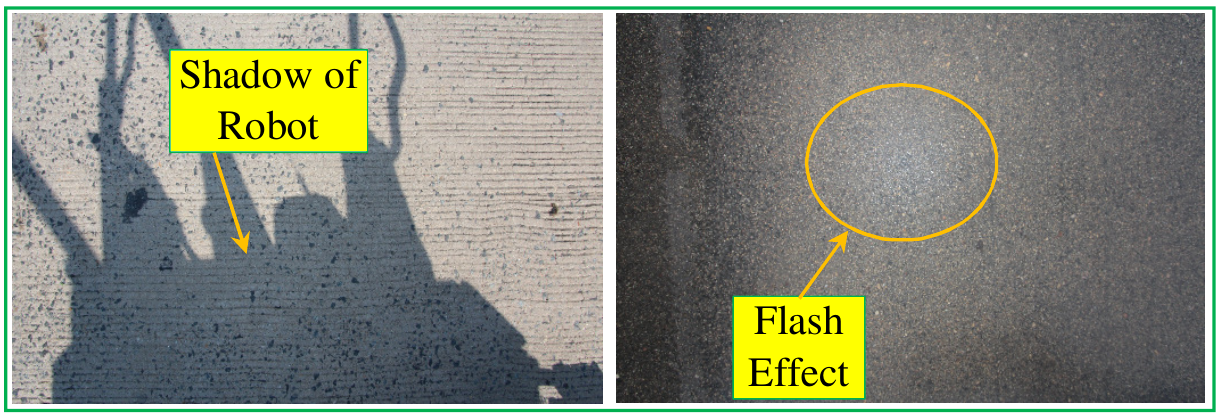}
\caption{(Left) Shadow problem with day time image collection. (Right) Flash light effect at nigh time image collection.}
\label{shadow effect}
\end{figure} 

Two wide-lens Cannon cameras are integrated with the robot. The robot collects images at every 2ft (0.61cm). Each of the cameras
covers an area of a size of 1.83m $\times$ 0.6m (Fig. \ref{image samples}). The images simultaneously
collected by these two cameras have about 30$\%$ overlap area that is used for image stitching. The automated image stitching algorithm was reported in the previous work \cite{La_Springer_2015}. Each camera is equipped with a flash to remove shadow at night time and mitigate shadow at day time (Fig. \ref{shadow effect}).
 
\begin{figure*}[htp!]
\centering
\includegraphics[width=16cm]{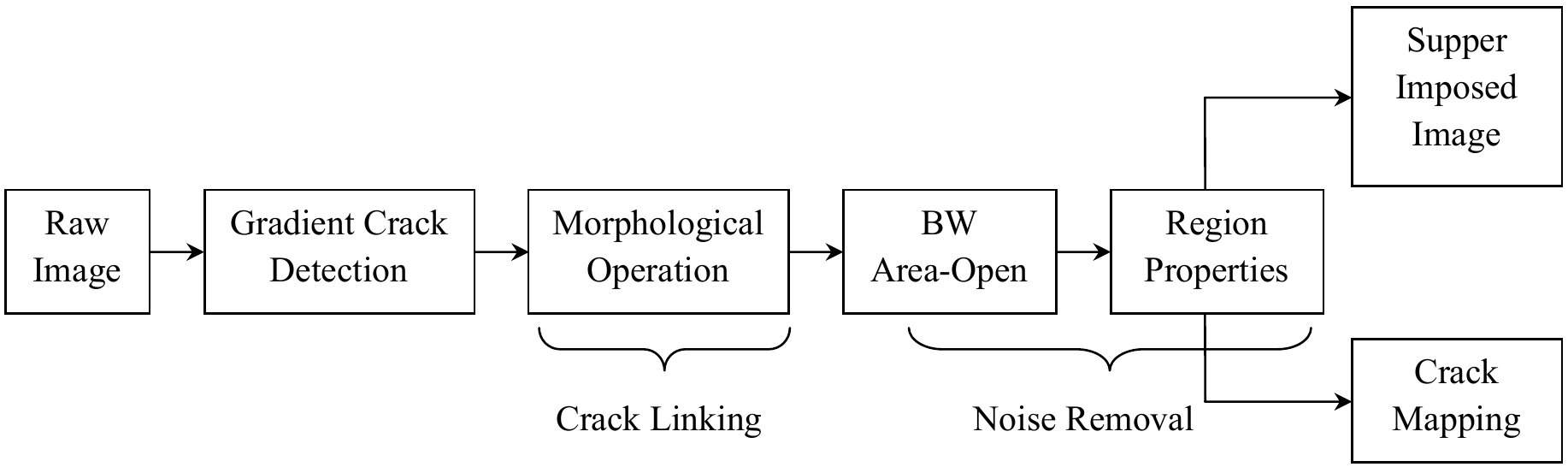}
\caption{Flowchart of the crack detection algorithm.}
\label{crack algorithm}
\end{figure*}

We propose a crack detection algorithm to detect crack on the stitched image with steps shown in Fig.~\ref{crack algorithm}. In the following, we describe three major steps in the algorithm: crack detection, crack linking and noise removal.

\subsubsection{Gradient crack/edge detection}
The goal of the crack detection module is to identify and quantify the possible crack pixels and their orientations. Let $I(x,y)$ be the source image at pixel $(x,y)$. We calculate the gradient vector of the intensity $I(x,y)$ as
\begin{equation}
\bs{I}_g=\nabla I(x,y) =  \frac{\partial I}{\partial x} \bs{i} +
\frac{\partial I}{\partial y} \bs{j}=I_x \bs{i} + I_y \bs{j},
\label{gradient}
\end{equation}
where $I_x=\frac{\partial I}{\partial x}$ and $I_y=\frac{\partial I}{\partial y}$ and $(\bs{i},\bs{j})$ are the gradient elements and unit vectors along the $x$- and $y$-axis directions, respectively.  We refine and extend the above gradient operator~(\ref{gradient}) by considering the edge/crack orientation in the diagonal directions besides the horizontal ($x$-axis) and vertical ($y$-axis) directions. We introduce eight gradient kernels to compute the gradients of $I(x,y)$. The eight $3\times 3$ convolution kernels $\bs{L}_\theta$, $\theta=\frac{\pi}{4}k$, $k=0,\cdots,7$, are defined in~(\ref{equ2}), and $\bs{L}_{\frac{4\pi}{4}}=\bs{L}^T_{\frac{2\pi}{4}}$, $\bs{L}_{\frac{5\pi}{4}}=\bs{L}^T_{\frac{\pi}{4}}$ and $\bs{L}_{\frac{6\pi}{4}}=\bs{L}^T_{\frac{2\pi}{4}}$. By calculating the convolution $I(x,y)\ast \bs{L}_\theta$, we obtain an approximation of the gradient/derivatives of the image intensity function (\ref{gradient}) along the orientation $\theta$. These kernels are applied separately to the input image, to produce separate measurements of the gradient component in each orientation. These calculations are also combined together to find the absolute magnitude of the gradient at each point and the orientation of that gradient.
\begin{figure*}[ht!]
\small
\begin{equation}
\text{\hspace{-2mm}} \bs{L}_0=\begin{bmatrix} 1 & 0 & -1 \\ 1 & 0 & -1 \\ 1 & 0 &
-1\end{bmatrix}, \bs{L}_{\frac{\pi}{4}}=\begin{bmatrix} 0 &
-1 & -1 \\ 1 & 0 & -1 \\ 1 & 1 & 0\end{bmatrix},
\bs{L}_{\frac{2\pi}{4}}=\begin{bmatrix} -1 & 
-1 & -1 \\ 0 & 0 & 0 \\ 1 & 1 & 1\end{bmatrix},
\bs{L}_{\frac{3\pi}{4}}=\begin{bmatrix} -1 & -1 & 0 \\ -1 & 0 & 1 \\ 0 & 1 &
1\end{bmatrix}, \bs{L}_{\frac{7\pi}{4}}=\begin{bmatrix} 1 & 1 & 0 \\ 1
& 0 & 1 \\ 0 & -1 & 
-1\end{bmatrix}, 
\label{equ2}
\end{equation}
\hrulefill
\vspace*{-2pt}
\end{figure*}

\subsubsection{Crack/edge cleaning and linking}

After applying the gradient crack detection process, a crack cleaning and linking process is applied to remove noise and link the crack pixels to form a continuous crack. Crack cleaning is performed via the Morphological operation~\cite{Jahan2013b}. These operations can remove isolated pixels and link pixels in the small neighborhood windows if most pixels in these windows are the crack pixels.

\begin{figure}
\centering
\includegraphics[width=7cm]{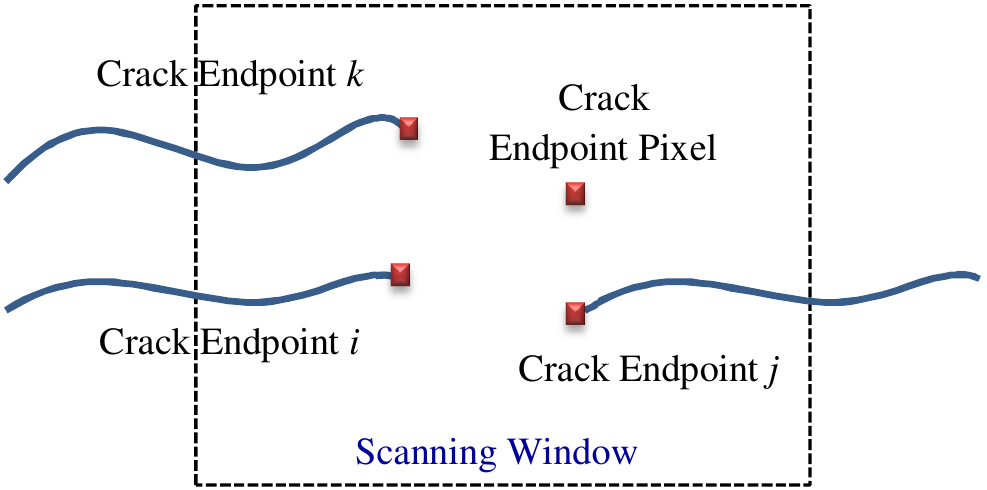}
\caption{Schematic of identifying the linking crack path to the $j$th crack path.}
\label{crack linking}
\end{figure}

In the crack linking process, the first step is to identify the starting and the ending points of the crack. Once this step is established, the crack linking process defines the scanning window size and then determines the maximum linking distance.  To decide the
linking direction, a cost function for the $j$th crack path is defined as 
\begin{equation}
F_{j}(i) = K_p\sqrt{(x^i_{ep} - x^j_{ep})^2+(y^i_{ep} - y^j_{ep})^2} +
K_d,
\label{Cost function}
\end{equation}
for any $i$th crack in the scanning window area, where $(x^i_{ep}, y^i_{ep})$ and $(x^j_{ep}, y^j_{ep})$ are the crack end-point locations of the $i$th and $j$th crack paths, respectively; see Fig.~\ref{crack linking}.
 Parameters $K_p$ and $K_d$ are constants that will be experimentally determined. After calculating $F_j(i)$ for all cracks in the scanning window, the minimum value of the cost function, $i^*=\arg \min_i F_j(i)$, determines the $i^*$th crack path to be linked to the $j$th crack path. 

\subsubsection{Noise removal}
We first remove small noisy connected components which have fewer than certain pixels in the area of two-dimensional eight-connected neighborhood. 

We  developed a process to further remove noise pixels by looking at crack area. Let $(x^i_c, y^i_c)$ be the center location of the $i$th crack region in the crack area. We first compute the distance among these crack regions as 
\begin{equation}
d(i,j) = \sqrt{(x^i_c - x^j_c)^2+(y^i_c - y^j_c)^2}, i,j = 1,\cdots,N_c,
\label{Dist}
\end{equation}
where $N_c$ is the total number of centroids of the crack regions.  We then combine the calculated distances $d(i,j)$ and the total areas $A(i)$ for the $i$th crack region to remove noises. The crack area $A(i)$ is calculated by the total number of pixels covered by detected $i$th crack region. With known $d(i,j)$ and $A(i)$, we compare their values with the predefined thresholds $T_d$ and $T_a$, respectively. If both the values of $d(i,j)$ and $A(i)$ are smaller than these thresholds, then $i$th crack region is removed from the detected candidate pool. 

\begin{figure}
	\centering
	\includegraphics[width=\columnwidth]{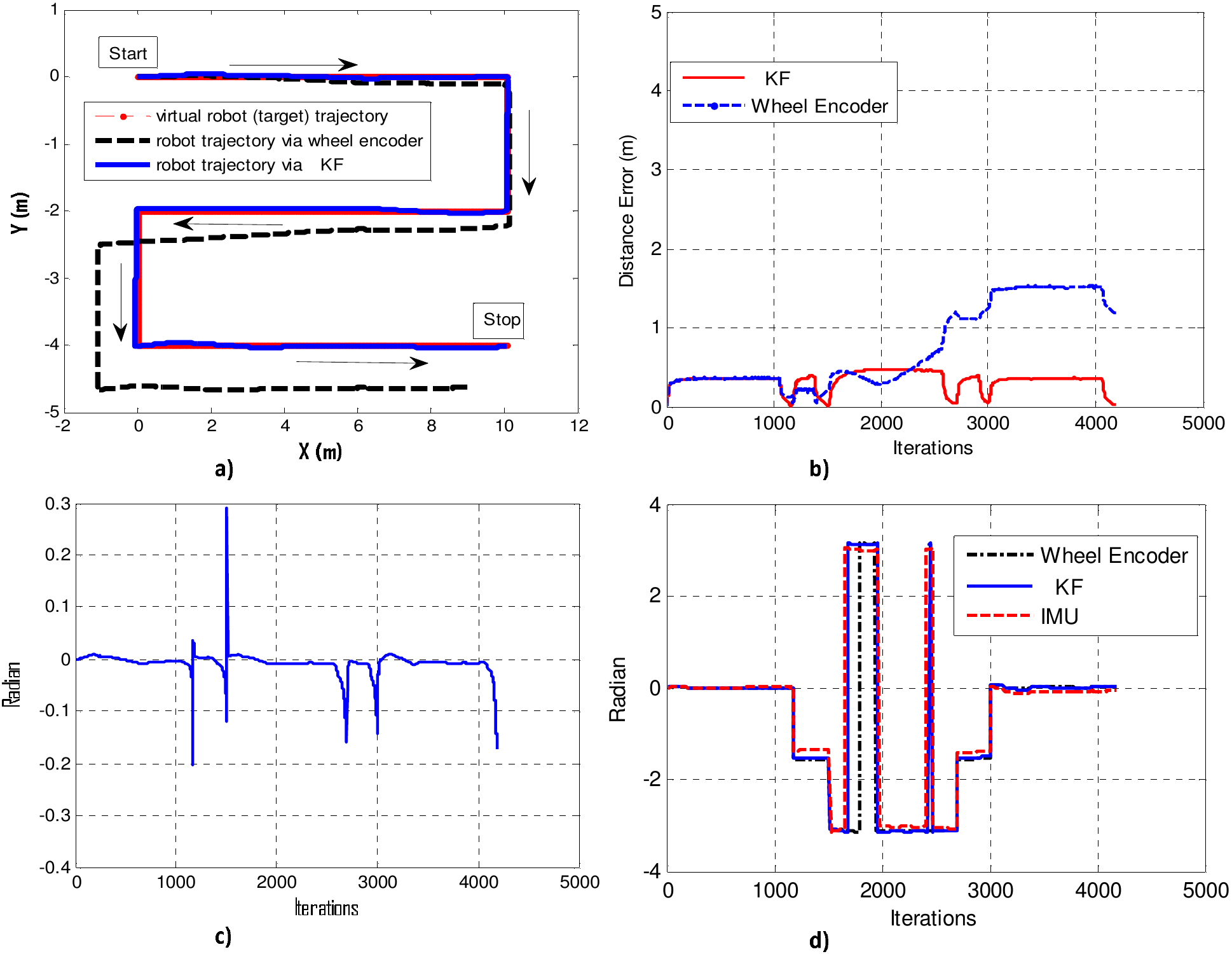}
	\caption{Experimental results of a straight bridge path planning.}
	\label{straight test}
\end{figure}
\section{Field Tests and Deployment Results}
\label{Results}
\subsection{Autonomous Navigation Tests}

\begin{figure}
	\centering
	\includegraphics[width=\columnwidth]{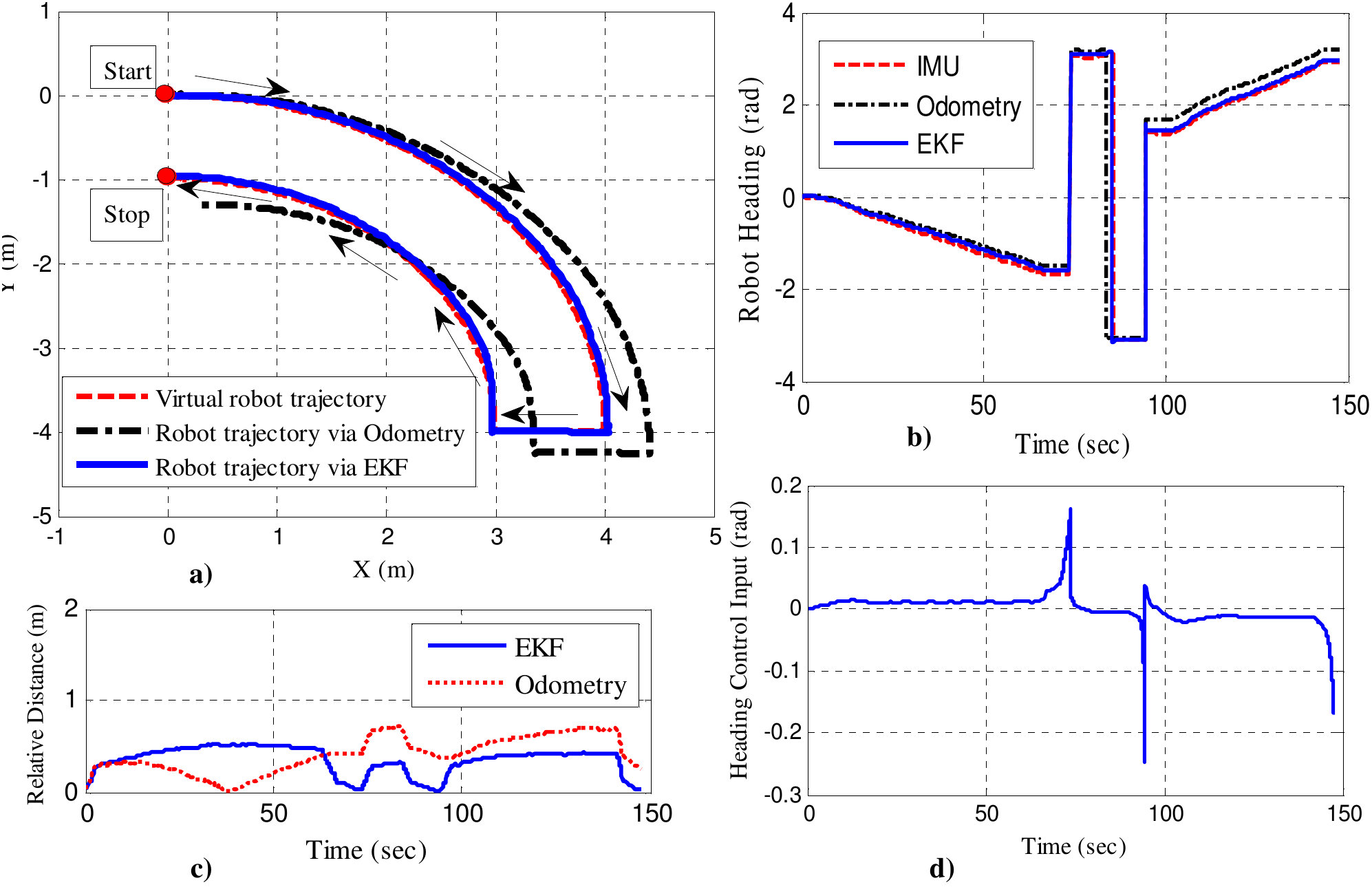}
	\caption{Experimental results of a curving bridge path planning.}
	\label{curving test}
\end{figure}

The autonomous navigation for the robot's path planning on the bridge is tested for both straight and curving bridge decks as shown in Fig. \ref{straight test} and \ref{curving test}. The comparison of the EKF-based localization and the odometry-only trajectory clearly demonstrates that the EKF-based localization outperforms the odometry-only trajectory.  The robot can follow the virtual one very well with the EKF localization, but large error with the odometry. For motion control performance, the virtual robot trajectory is plotted in Fig.  \ref{straight test}-a and  $\ref{curving test}$-a, and we can see that the robot follows the virtual robot closely. To further demonstrate the navigation performance, Fig.  \ref{straight test}-b and  $\ref{curving test}$-c show the comparison results of the relative distance between the robot and the virtual one among the navigation based EKF and odometry, respectively.  Fig. \ref{straight test}-d and \ref{curving test}-b show the robot heading obtained from IMU, Odometry and EKF, respectively during the tracking process.  Fig. \ref{straight test}-c and $\ref{curving test}$-d show the heading control input applying to the robot. 

\subsection{Deployment Results}
\begin{figure}[htp!]
\centering
\includegraphics[width=12cm]{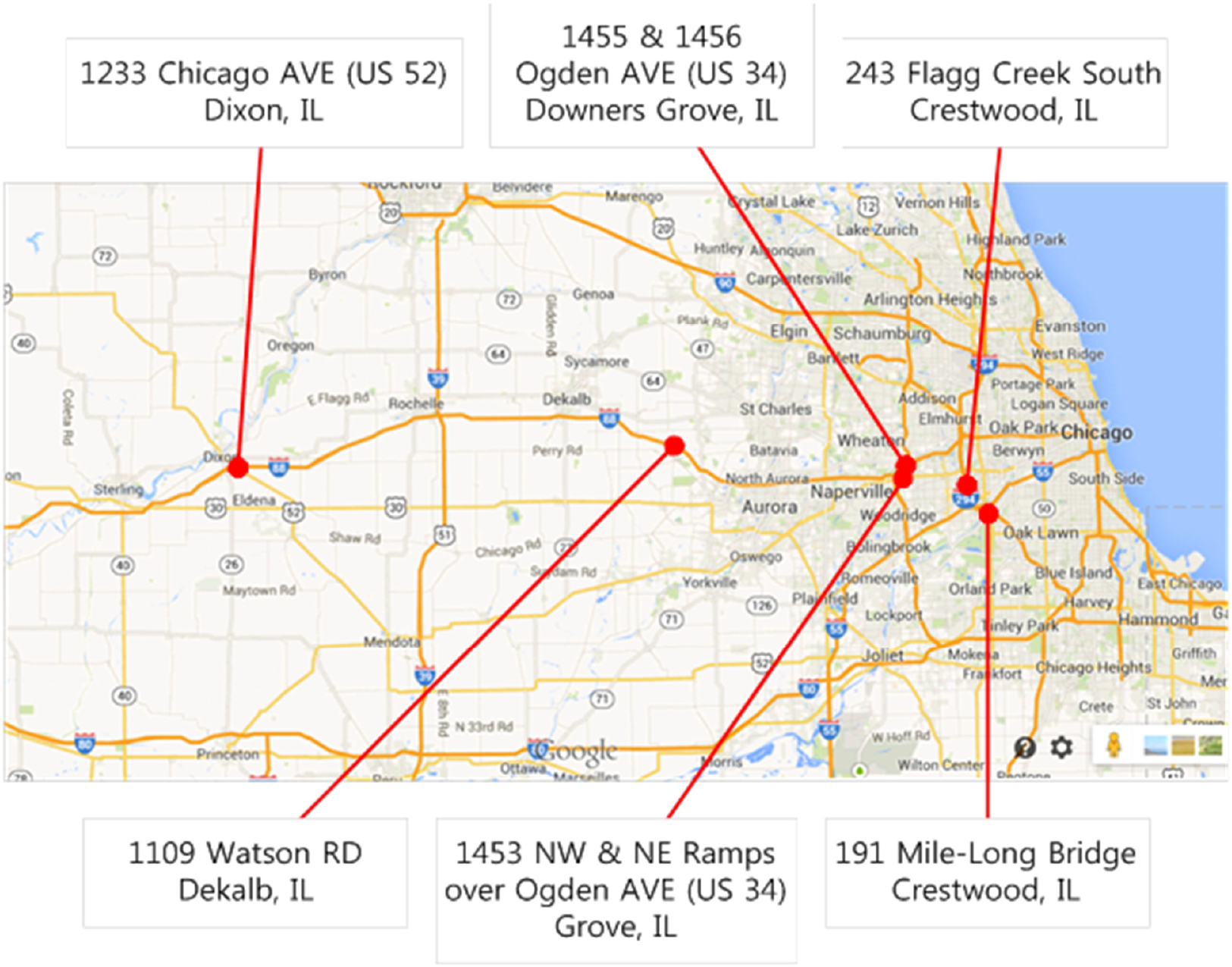}
\caption{Location of bridges surveyed by the developed robot in Illinois in December 2013 and April 2014.}
\label{test locations}
\end{figure}

\begin{figure*}[htp!]
\centering
\includegraphics[width=\textwidth]{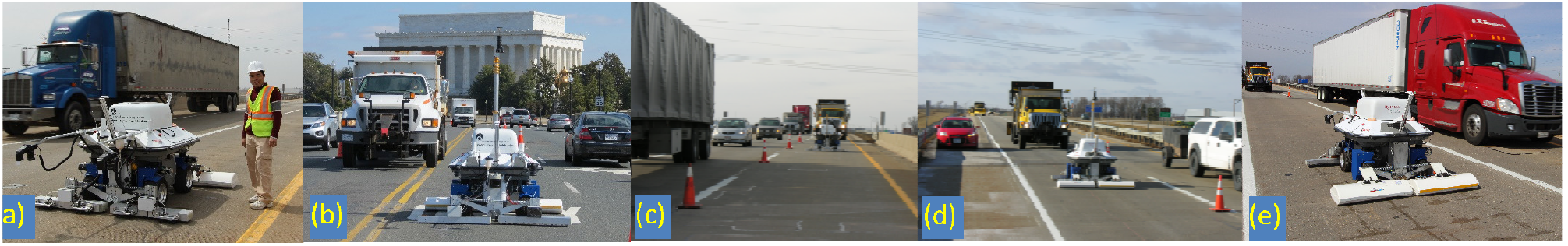}
\caption{Deployment of the NDE robot for inspection of more than 40 bridges: (a) Inspect the Chicago Mile-long bridge in Illinois April 2014 (the co-author Hung La with the robot in this picture); (b) Inspect the Arlington memorial bridge in DC, April 2014; (c) Inspect the Haymarket highway bridge in Virginia, May 2014; (d) Inspect the Ogden avenue bridge, Downers Grove, Illinois, December 2014; (e) Inspect the Chicago avenue bridge in Dixon, Illinois in April 2014.}
\label{RABIT_bridge}
\end{figure*}

The proposed robotic system has been deployed on more than  40 bridges in New Jersey, Virginia, Washington DC, Maryland, Pennsylvania, Illinois, etc. as reported in \cite{La_NJ_report2014, La_Chicago_report2014}. For example, as shown in Fig. \ref{test locations}, the robot was deployed in several different places in nearby Chicago, Illinois in December 2013 and April 2014 to inspect bridges there. Fig. \ref{RABIT_bridge} shows the robot inspecting five different highway bridges in States of Illinois and Virginia, respectively. Due to similar representation of the system deployments and inspection results, we just present the results of two example bridges: Haymarket highway bridge in Virginia and Ogden avenue bridge in Illinois, USA.

\begin{figure*}[htp!]
\centering
\includegraphics[width=\textwidth]{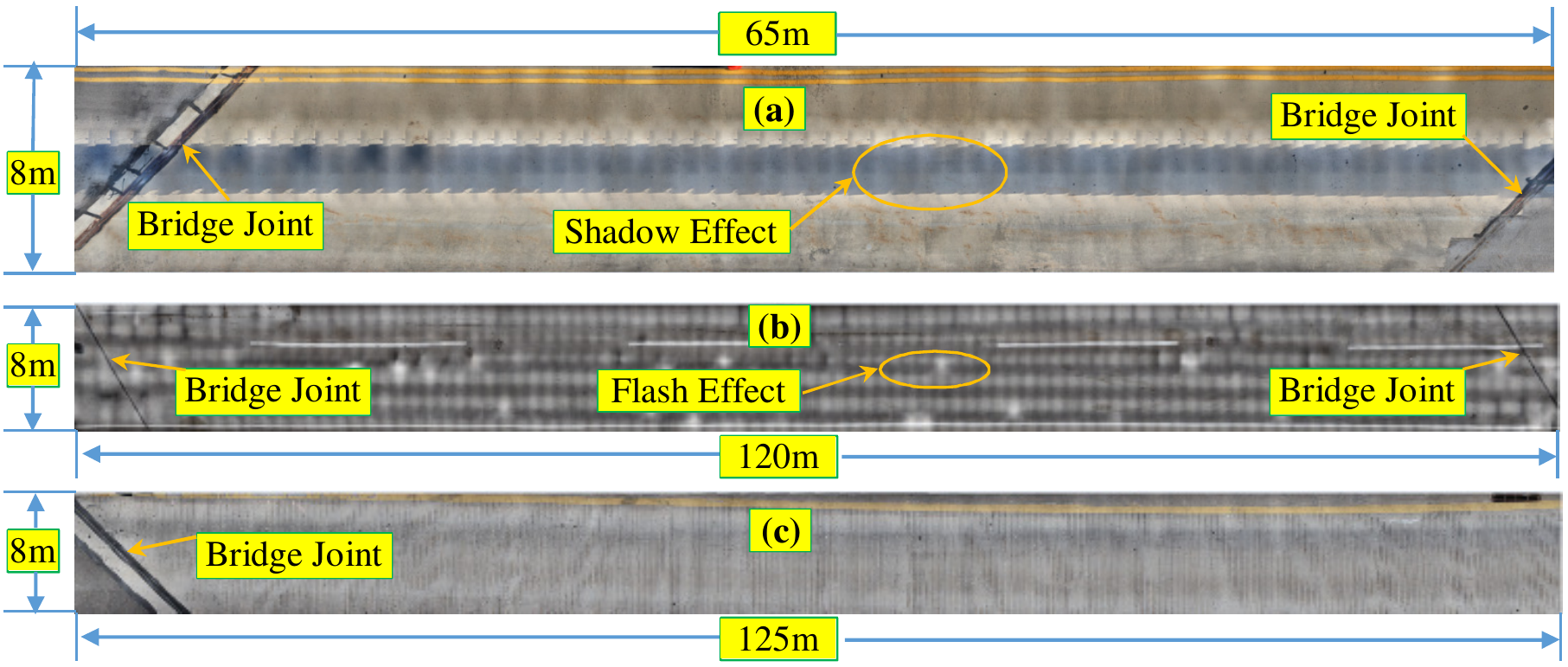}
\caption{Deck image of three different bridges in different scenarios: (a) stitched image deck from shadowed individual images; stitched deck image from individual images collected at night time; (a) stitched deck image from individual images collected at day time but no shadow.}
\label{stitching result}
\end{figure*}
\begin{figure*}[htp!]
\centering
\subfigure[]{
\label{crack image result}
\includegraphics[width=\textwidth]{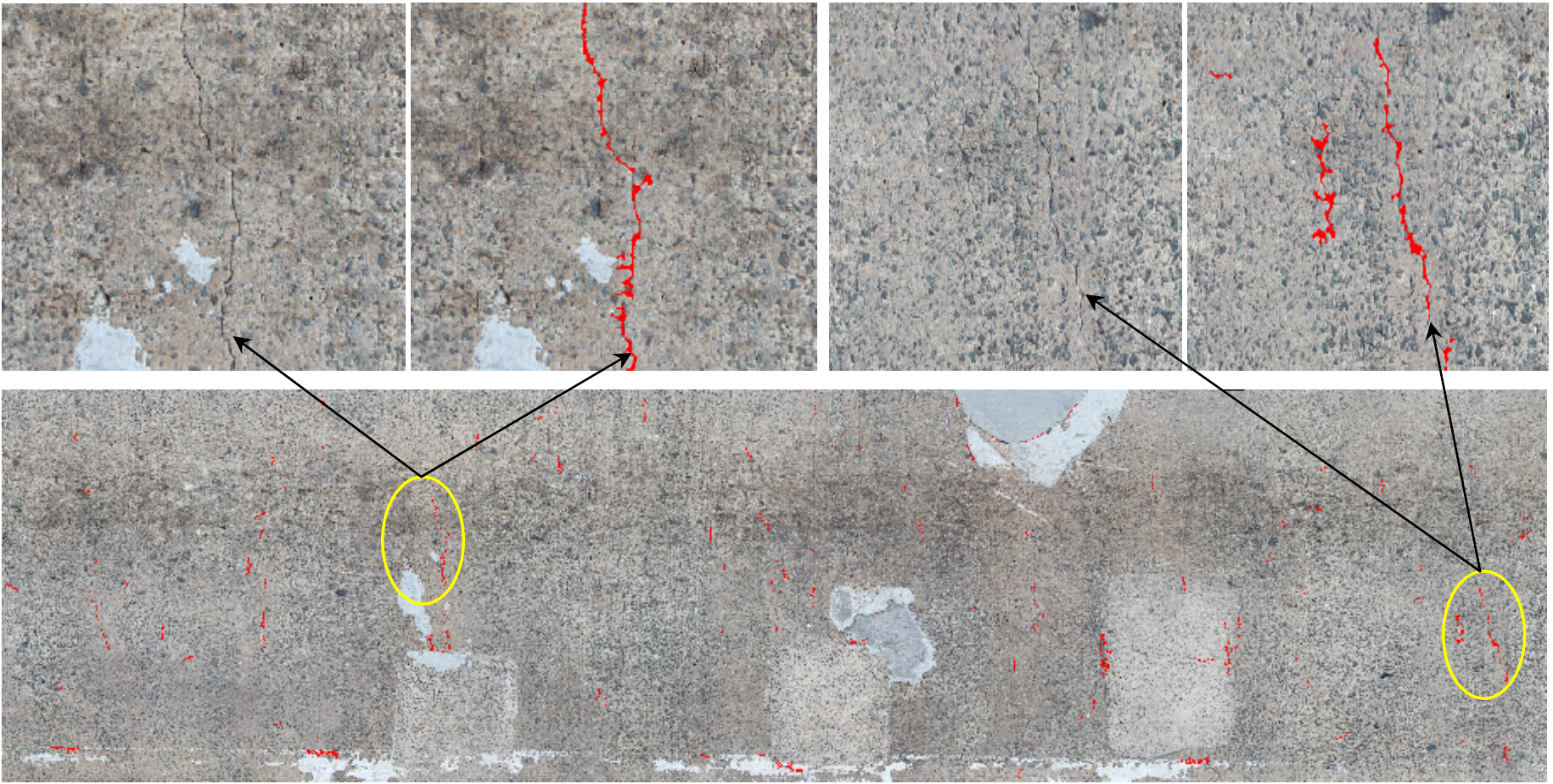}}
\subfigure[]{
\label{crack location result}
\includegraphics[width=\textwidth]{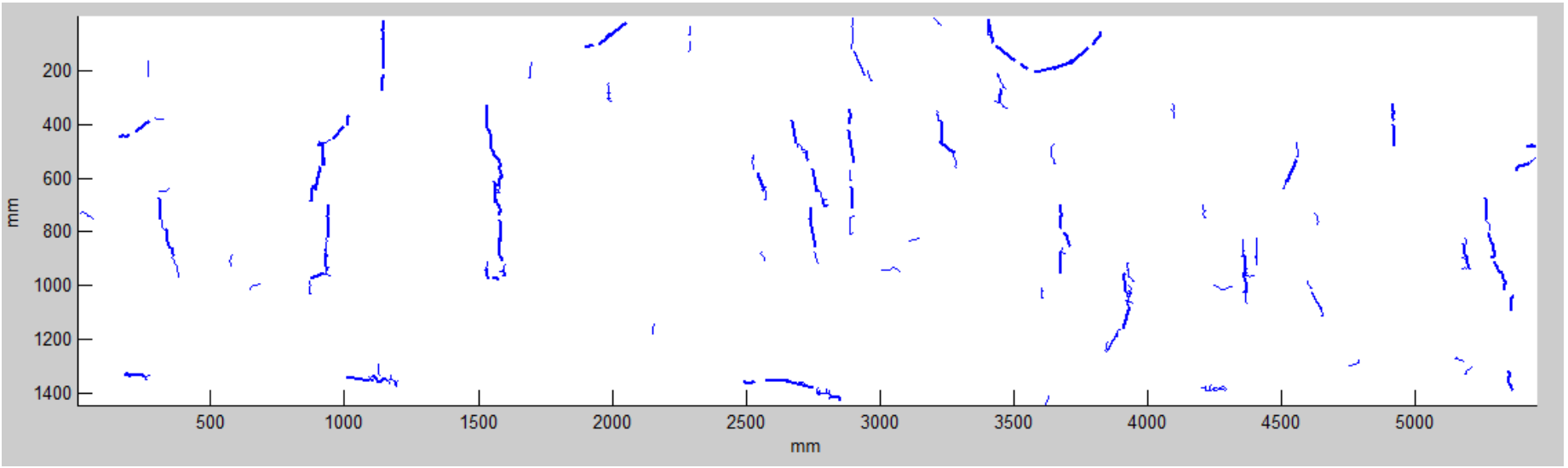}
}
\caption{Crack detection and mapping results for a bridge deck area on the Haymarket bridge in Virginia in June 2013. (a) Stitched image from robotic collected images with several crack samples. (b) Crack map with crack sizes and lengths.}
\end{figure*}

\subsubsection{Visual Data Analysis}

For automated assessment of the deck, all individual images are stitched together to present the entire bridge deck picture as shown in Fig. \ref{stitching result}. For more information of the image stitching algorithm please refer to  \cite{La_ISARC_2014, La_Springer_2015}. The crack detection is performed in the stitched image of Haymarket bridge over an area of a size of $5.5m \times 1.45m$.  The top sub-figures are the zoom-in images at several crack locations for clear presentation and demonstration. Fig.~\ref{crack location result} shows the crack detection and mapping results for the same bridge deck area shown in Fig.~\ref{crack image result}.

To provide more details of bridge deck visual condition, the crack detection and mapping results not only localize the cracks on the bridge but also provide the sizes of these cracks. Table~\ref{Crack Stat} lists the statistics of the detected cracks on the bridge deck area. From these calculations, we automatically obtain various statistical and location information about the cracks and their properties, which are both critical for quality assessment of the bridge deck.
\renewcommand{\arraystretch}{1.4}
\begin{table}[htb!]
\begin{center}
\setlength{\tabcolsep}{0.03in}
\caption{Statistics of the detected cracks for the bridge deck area shown in
  Fig.~\ref{crack image result}}   
\label{Crack Stat} 
  \begin{tabular}{|c|c|c|c|c|c|}
	   \hline\hline 
& Total & Longest & Shortest & Max. width & Min. width \\
	   \hline 
 Length& $10.6$ m& $70$ cm & $5$ cm & $1.5$ cm & $1.5$ mm \\
	   \hline 
Loc. $(x,y)$ (m) &  N/A & $(1.6, 0.7)$& $(5.5, 1.22)$ & $(5.45, 1.1)$
& $(4.2, 7.1)$ \\  \hline\hline  
  \end{tabular} 
\end{center}
\end{table}
\begin{figure}[htp!]
\centering
\includegraphics[width=12cm]{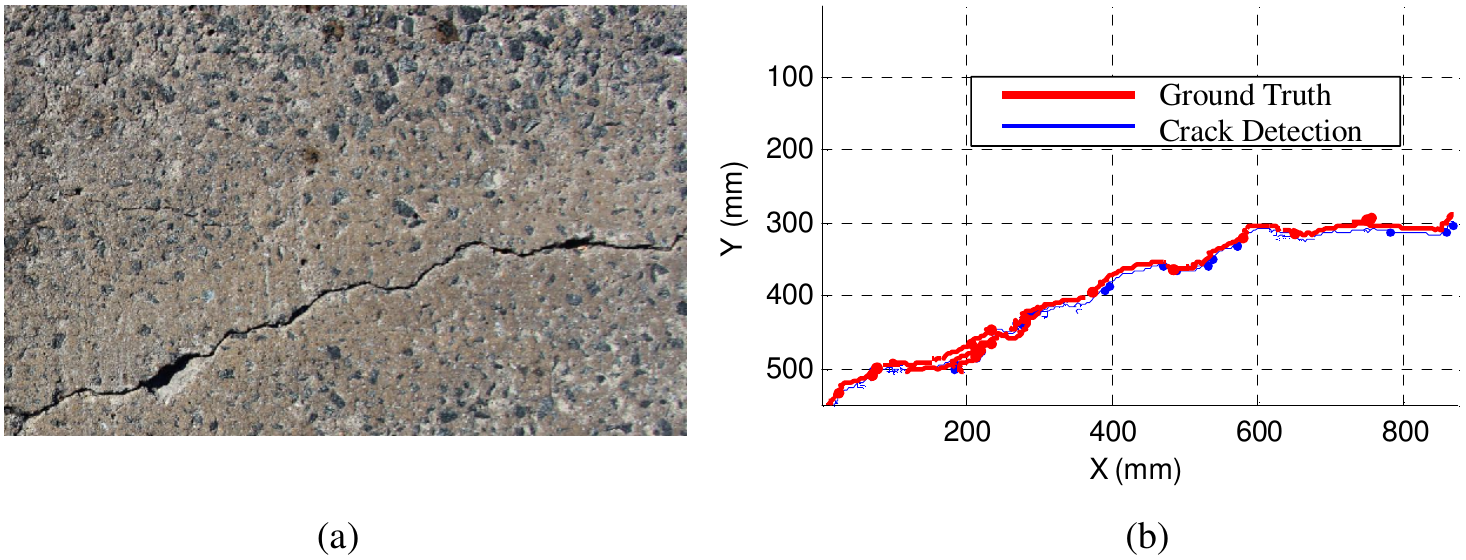}
\caption{(a) Crack on an concrete surface. (b) Comparison between ground truth crack extracted by the inspector and the crack detection result by the proposed algorithm.} 
\label{Comparison}
\end{figure}

Fig.~\ref{Comparison} shows the comparison results of the detected
cracks with the ground truth. The image shown in
Fig.~\ref{Comparison}(a) was taken at the concrete bridge deck at
Scott Street, Arlington, VA. The crack was first inspected and
extracted by the bridge inspector. We consider the crack image
collected by the human inspector as the ground truth. We then applied
our crack detection algorithm to detect cracks on this image and
overlaid the crack detection results with the ground truth as shown in
Fig.~\ref{Comparison}(b). These results can show the accuracy of the crack detection algorithm on real field collected
bridge deck images.

We also compare the proposed crack detection algorithm with four other
commonly used crack/edge detection methods \cite{Mohammad_SIE_2009}: Laplacian of
Gaussian (LoG) method~\cite{Forstyth}, Canny edge detection method~\cite{Canny1986}, Haar Wavelet edge detection method \cite{Bachman_NY_2010, Qader_JCCE_2003}, and Percolation-based method \cite{Yamaguchi_MVA_2010}. The LoG method combines Gaussian filtering with the Laplacian for edge detection. The Canny edge detector uses linear filtering with a Gaussian kernel to smooth the noise in the image.  Haar Wavelets edge detection method \cite{Bachman_NY_2010, Qader_JCCE_2003} based on the fast Haar transform (FHT) to decompose the image into low-frequency and high-frequency components. Then, the process isolates those high-frequency coefficients from which the edge features of an image are identified. 
Percolation-based crack detection \cite{Yamaguchi_MVA_2010} is a type of scalable local processing method that considers
the connectivity among neighboring pixels. This method enables the evaluation of whether or not the central pixel in a local window is a crack based on the shape of the cluster formed by the percolation processes. 

\begin{figure}[htp!]
\vspace{-15pt}
\centering
\includegraphics[width=17cm, height =13cm]{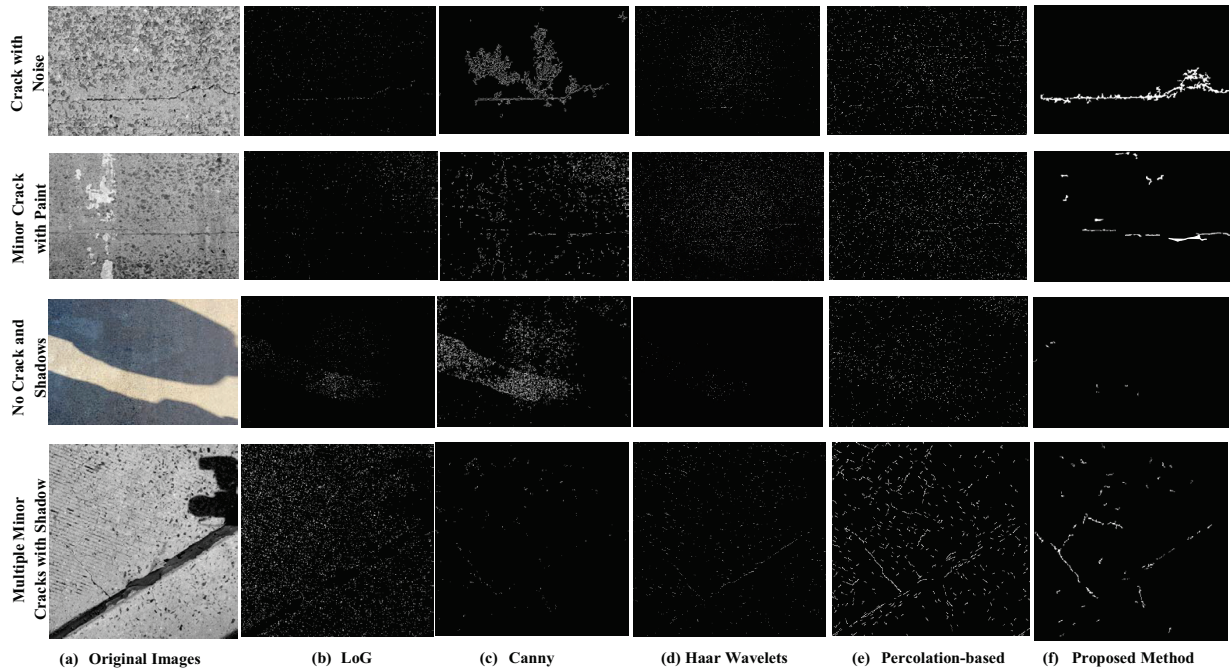}
\vspace{-15pt}
\caption{Comparison among various types of bridge deck images. (a) Original
  concrete bridge deck image. (b) LoG edge detection results. (c)
  Canny edge detection results. (d) Harr Wavelets crack detection. (e) Percolation-based crack detection. (f) The proposed crack detection
  results.}
\label{crack compare}
\end{figure}

We implemented and compared the crack detection
methods on different concrete deck images: image with noise, image
with paint, image with shadows, and image with small crack
sizes. 
Fig.~\ref{crack compare} shows the comparison results among LoG, Canny, Harr Wavelets edge detection methods, percolation-based crack detection method, and our crack detection method. When
implementing the LoG, Canny and Harr Wavelets and Percolation-based methods, respectively, we selected the
parameters to obtain the best crack detection results. As can be seen in Fig.~\ref{crack compare}, the Harr Wavelets method performed better than LoG and Canny since it can detect cracks with less noise than the latter ones. This is consistent with the findings reported in \cite{Qader_JCCE_2003, Mohammad_SIE2009}. We can also see that the recent developed Percolation-based method \cite{Yamaguchi_MVA_2010} performed well in detecting cracks in images with less noisy background (Fig.~\ref{crack compare}e-Bottom), but fail with images has more noisy background (Fig.~\ref{crack compare}e-Top). The results of the proposed crack detection method are shown in (Fig.~\ref{crack compare}f) which clearly demonstrates that our method outperforms the other ones.

\subsubsection{NDE Data Analysis}

\begin{figure*}[htb!]
\centering
\includegraphics[width=11cm, angle =-90]{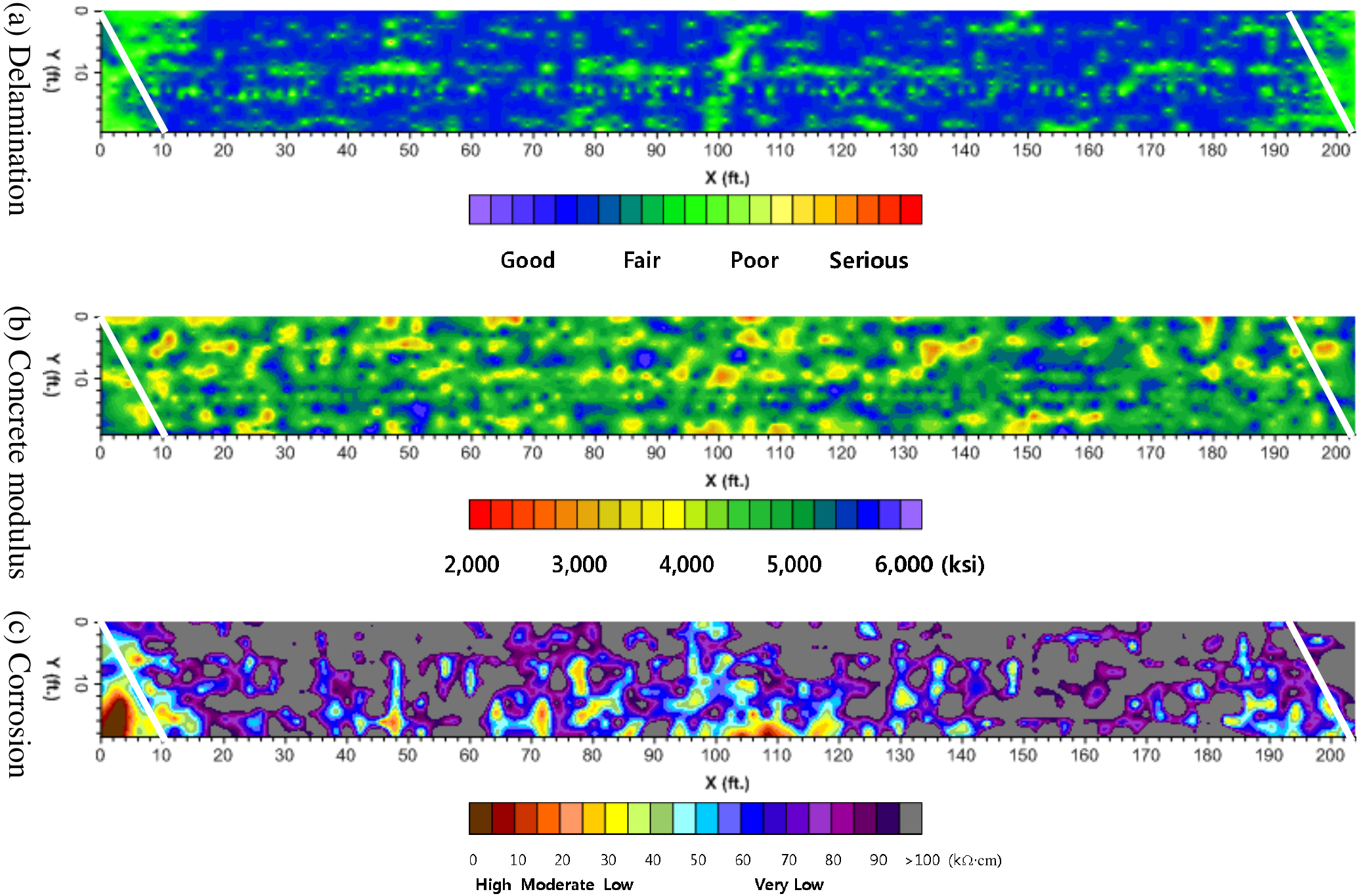}
\caption{(a) Impact-echo (IE) delamination; (b) Ultrasonic surface waves (USW) elastic modulus; and (c) Electrical resistivity (ER) corrosion condition maps of the Ogden avenue bridge deck in Illinois (April 2014) based on data collected by the developed robot system. The robot is autonomously maneuvered  on the deck based on the proposed linear and omni navigation algorithms (\ref{nav_algorithm}) and (\ref{map_xy}). The robot covers the bridge deck area of 20ft$\times$200ft (6.1m$\times$61m) within 40 minutes.}
\label{IE_USW_fig}
\end{figure*}

Two acoustic arrays are integrated with the robot and each array can produce 8 Impact-Echo (IE) and 6 Ultrasonic Surface Waves (USW) data set at each time of measurment.  IE is an elastic-wave based method to identify and characterize delaminations in concrete structures.  This method uses the transient vibration response of a plate-like structure subjected to a mechanical impact.  The mechanical impact generates body waves (P-waves or longitudinal waves and S-waves or transverse waves), and surface-guided waves (e.g. Lamb and Rayleigh surface waves) that propagate in the plate.  In practice, the transient time response of the solid structure is commonly measured with a contact sensor (e.g., a displacement sensor or accelerometer) coupled to the surface close to the impact source. The fast Fourier transform (amplitude spectrum) of the measured transient time-signal shows maxima (peaks) at certain frequencies, which represent particular resonance modes. To interprete the severity of the delamination in a concrete deck with the IE method, a test point is described as solid if the dominant frequency corresponds to the thickness stretch modes (Lamb waves) family. In that case, the frequency of the fundamental thickness stretch mode (the zero-group-velocity frequency of the first symmetric ($S_1$) Lamb mode, or also called the IE frequency ($f_{IE}$). The frequency can be related to the thickness of a plate $H$ for a known $P$-wave velocity $C_p$ of concrete by 
\begin{equation}
H = \frac{\beta_1 C_p}{f_{IE}}
\label{IE_calculate}
\end{equation}                                                 
where $\beta_1$ is a correction factor that depends on Poisson's ratio of concrete, ranging from 0.945 to 0.957 for the normal range of concrete. A delaminated point in the deck will theoretically demonstrate a shift in the thickness stretch mode toward higher values because the wave reflections occur at shallower depths. Depending on the extent and continuity of the delamination, the partitioning of the wave energy reflected from the bottom of the deck and the delamination may vary. Progressed delamination is characterized by a single peak at a frequency corresponding to the depth of the delamination. In case of wide or shallow delaminations, the dominant response of the deck to an impact is characterized by a low frequency response of flexural-mode oscillations of the upper delaminated portion of the deck.  The IE delamination condition map is presented in Fig. \ref{IE_USW_fig}-a.  Hot colors (reds and yellows) are an indicator of delamination, while cold colors (greens and blues) are an indicator of likely fair or good conditions. As can be observed, the deck is generally in a good condition (less delamination), with some signs of incipient/initial delamination indicated by green, and very few signs of progressed delamination indicated by yellow colors. There were only a few points where the delamination was identified as fully developed (red colors).

The USW technique is the spectral analysis of surface waves (SASW) to evaluate material properties (elastic moduli) in the near-surface area The SASW uses the phenomenon of surface wave dispersion (i.e., velocity of propagation as a function of frequency and wave length, in layered systems to obtain the information about layer thickness and elastic moduli). A SASW test consists of recording the response of the deck, at two receiver locations, to an impact on the surface of the deck.  The surface wave velocity can be obtained by measuring the phase difference  $\Delta \phi$ between two different sensors as
\begin{equation}
C = 2\pi f\frac{d}{\Delta \phi}
\label{USW_calculate}
\end{equation}
where $f$ is frequency, $d$ is distance between two sensors. The USW test is identical to the SASW, except that the frequency range of interest is limited to a narrow high-frequency range in which the surface wave penetration depth does not exceed the thickness of the tested object.  Significant variation in the phase velocity will be an indication of the presence of a delamination or other anomaly.  The concrete modulus USW map, as shown in the map in Fig. \ref{IE_USW_fig}-b, varies between about 2000 and 6000 $ksi$. The test regions classified as serious condition (red color) are interpreted as likely delaminated areas on the concrete deck. The USW map  provides the condition assessment and quality of concrete through measuring concrete modulus. 

Four ER probes  are integrated with the robot to evaluate the corrosive environment of concrete and thus potential for corrosion of reinforcing steel. Dry concrete will pose a high resistance to the passage of current, and thus will be unable to support ionic flow. On the other hand, presence of water and chlorides in concrete, and increased porosity due to damage and cracks, will increase ion flow, and thus reduce resistivity. Research has shown in a number of cases that ER of concrete can be related to the corrosion rates of reinforcing steel. The ER surveys are commonly conducted using a four-electrode Wenner probe, as illustrated in Fig. \ref{NDE sensors}-b. Electrical current is applied through two outer electrodes, while the potential of the generated electrical field is measured using two inner electrodes, and from the two, ER is calculated. The robot carries four electrode Wenner probes and collects data at every  two feet (60cm) on the deck. To create a conducted environment between the ER probe and the concrete deck, the robot is integrated with the water tank and to spray water on the target locations before deploying the ER probes for measurements. The concrete corrosion ER map is shown in Fig. \ref{IE_USW_fig}-c.  Overall, good condition of the corrosion rate was identified throughout the scanned bridge deck since only some small areas have  moderate  to  high corrosion rate (red color).

\begin{figure*}[htb!]
\centering
\vspace{-60pt}
\includegraphics[width=16cm]{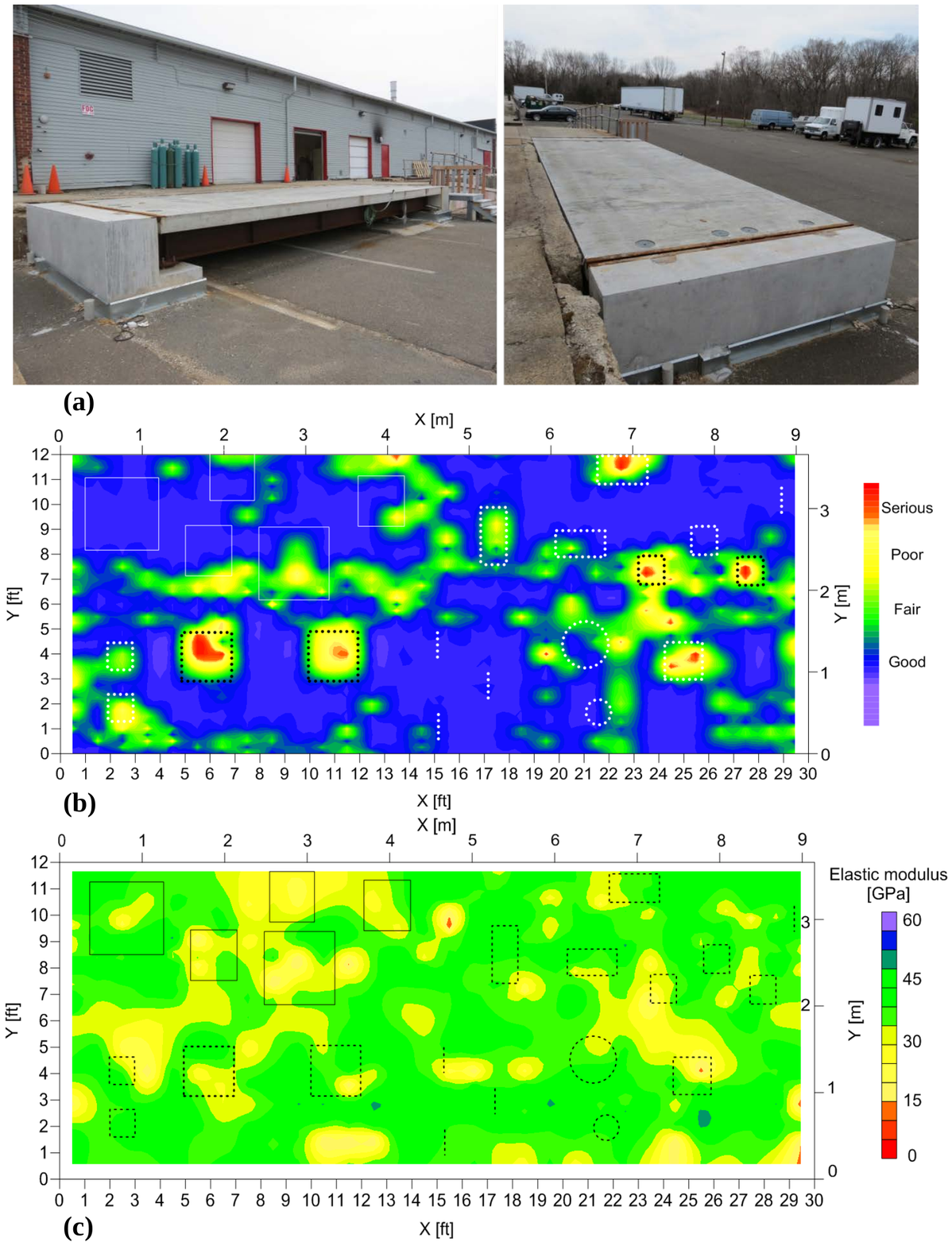}
\vspace{-60pt}
\caption{(a) Concrete slab with artificial defects for validation; (b) Delaimination condition map of the concrete slab using IE test; (c) Elastic modulus condition map of the concrete slab using USW test.  The squares/rectangles and circles are artificial defect areas (ground truth).}
\label{NDE validation}
\vspace{-0pt}
\end{figure*}

To evaluate above mentioned NDE technologies, the simulated deck was built as a ground truth as shown in Fig.  \ref{NDE validation}(a) which contains four different types of artificial defects: delaminations having various depths and areal extent, surface-breaking cracks having various depths, deteriorated regions with reduced elastic modulus, and four cable conduits (three of zinc and one of plastic) including steel strands with different grouting conditions. The delaminations were fabricated by using two layers of plastic foam pieces covered by thin plastic film with various sizes and at three different depths. Shallow delaminations was placed at a 5 cm (2 in.) depth, intermediate delaminations at a 10 cm (4 in.) depth, and deep delaminations at a 17 cm (6.5 in.) depth. To ensure that the delaminations were positioned at the designed depth, the thickness of the slab was divided into four layers (a layer including the bottom reinforcing steel, middle of the slab, a layer including the top reinforcing steel, and the top surface of the slab), and concrete was cast layer by layer. Surface-breaking cracks were built in the deck by inserting a layer of plastic sheets before concrete casting. The design depths of the four vertical cracks were 2.5, 5, 7.5 and 10 cm (1, 2, 3, and 4 in.), respectively. The deteriorated regions with reduced elastic modulus were prepared by inserting concrete blocks of a segregated or uniform size coarse aggregate.   In addition, about 25\% of the concrete deck was prepared for monitoring of chloride-induced deterioration through accelerated corrosion. Pockets of high chloride mix (15\% of Cl- by weight) was placed on the five selected regions during casting. Consequently, 45 kg of natural sea salt was uniformly distributed over the regions and mixed during concrete pouring.

We conducted validation of the IE and USW methods. Fig.  \ref{NDE validation}(b) is the delamination map of the concrete slab in Fig. \ref{NDE validation}(a) based on the IE method procedure. The resulting condition map confirms that IE method is effective in detecting and characterizing the most of the delaminations in the concrete deck. The locations of shallow delaminations shown as red spots or areas, indicating ``serious condition" in the delamination map. Deep and intermediate delaminations are shown as green to yellow areas, indicating ``fair to poor condition" respectively. The locations of the areas with a reduced concrete modulus and with hollow or partially grouted ducts are also shown in the IE condition map. It can be seen that some of the artificial defects were missed in the condition map, primarily due to the lower spatial resolution. For the accelerated corrosion test region, the condition is marked as green to yellow, or ``fair to poor" condition. The peak frequencies obtained in those regions are only slightly lower than IE frequency for the solid regions. It can be physically interpreted that there is higher porosity and/or that micro cracks in concrete are developing due to corrosion activity in the salt contaminated test region, but have not caused delamination yet.

Fig. \ref{NDE validation}(c) is the concrete quality (modulus) map of the concrete slab in Fig. \ref{NDE validation}(a) based on the USW method procedure. The resulting modulus in the whole concrete deck ranges between 7.0 to 52.2 GPa with an average of 38.04 GPa and standard deviation of 8.46 GPa. Thus the coefficient of variation was about 22\%. The USW does point to the locations of artificial defects but lower accuracy than the IE due to the physical principle of the USW measurement and lower spatial resolution of the USW test setup than that of the IE test setup. However, the USW condition map provides a reasonably good correlation to the IE condition map in the accelerated corrosion test region. As the corrosion activity has influenced the P-wave velocity and, thus, the dominant frequency response in the IE test, it has reduced the velocity of surface waves in the USW test. 

\section{Conclusions}
\label{Con}

We have reported a new development of the autonomous mobile robotic system for bridge deck data collection, inspection and evaluation. Extensive testings and deployments of the proposed robotic system on many bridges proved the advantages and efficiency of the new automated nondestructive evaluation (NDE) approach for bridge deck inspection and evaluation. The accurate and reliable robotic navigation in the open-traffic bridge deck inspection is developed. Data collection and analysis for bridge deck crack detection, delamination (hidden crack) with IE, concrete modulus with USW, and concrete corrosion with ER have been presented. In all bridge inspection deployments, the robot was able to accurately and safely localize and navigate on the bridge decks to collect the NDE data.  The developed crack detection algorithm can detect cracks accurately in high noisy concrete images and build the whole crack map of the deck. The delamination, elastic modulus and corrosion maps were built based on the analysis of IE, USW and ER data collected by the robot to provide the ease of evaluation and monitor of the bridge. 

In the future work we will focus on the development of  sensor fusion algorithms for the NDE sensors and camera/visual data for a more comprehensive and intuitive bridge deck condition assessment and data representation. The developed robotic platform provides different types of data including visual data as well as multiple NDE sensory data from GPR, IE, USW and ER.  Efficient analysis and combination/fuse of these large amount of data  are challenging tasks. We plan to fuse and integrate the complementary NDE sensory data as through a probabilistic modeling framework. First, all defect information such as deterioration, concrete modulus, delamination, corrosion and local cracks will be obtained from each individual NDE sensor and visual cameras. Then, depending on the historical testing data and sensor models, a probabilistic fusion scheme will be built to construct the correlation model among these NDE sensors.

\newpage
\begin{center}
\textsc{APPENDIX}
\end{center}
In this appendix we present the proof of Theorem 1.

\textit{Proof:}

We choose a Lyapunov function as follows:
\begin{equation}
	   L= V_{a}=\frac{1}{2}\lambda q^{T}_{rv}q_{rv}=\frac{1}{2}\lambda \|q_{rv}\|^{2}.
		\label{Lyapunov}
\end{equation}
This function is positive definite, and the derivative of $L$ is given by
\begin{equation}
\dot{L}  = \frac{\partial L}{\partial q_{rv}} \dot{q}_{rv} = \frac{\partial L}{\partial q_{rv}} p_{rv}, 
\label{Lyapunov derivative one sensor}
\end{equation}
where the relative velocity between the mobile robot and the virtual robot is designed following the direction of negative gradient of $V_a$ with respect to $q_{rv}$ as:
\begin{equation}
  p_{rv}=-\nabla_{q_{rv}}V_{a} = -\nabla_{q_{rv}}L.
\label{vel}
\end{equation}
  Hence, substituting $p_{rv}$ given by (\ref{vel}) into ($\ref{Lyapunov derivative one sensor}$) we obtain
\begin{eqnarray}
\dot{L}   & = & 
   -\frac{\partial L}{\partial q_{rv}}\nabla_{q_{rv}}L = -\lambda^{2} \|q_{rv}\|^{2}\nonumber \\ 
		&& =
		 -2\lambda \frac{1}{2}\lambda \|q_{rv}\|^{2} = -2\lambda	V_{a}<0.
\label{Lyapunov derivative2}
\end{eqnarray}

From (\ref{Lyapunov}) and ($\ref{Lyapunov derivative2}$) we obtain
\begin{equation}
{\dot{V_{a}}} =  -2\lambda	V_{a}.
\label{Lyapunov derivative3}
\end{equation}
Solving this equation we get the solution as follows:
\begin{equation}
V_{a} =  V_{a}(0)e^{-2\lambda t}
\label{Lyapunov derivative3 one sensor}
\end{equation}
here $V_{a}(0)$ is the value of $V_{a}$ at $t=0$. This solution shows that $V_{a}$ and $\|q_{rv}\|$ converge to zero with the converging rate $\lambda$, or the position and velocity of the mobile robot asymptotically converges to those of the virtual robot after a certain time ($t>0$).

\section*{Acknowledgment}
The authors would like to thank Profs. Basily Basily and Ali Maher of Rutgers University for their support for the project development. The authors would also like to thank Spencer Gibb from Advanced Robotics and Automation Lab of University of Nevada, Reno for his support of implementing  Haar Wavelet edge detection and Percolation-based method for crack detection. The authors are also grateful to Ronny Lim, Hooman Parvardeh, Kenneth Lee and Prateek Prasanna of Rutgers University for their help during the system development  and field testing. 

\bibliographystyle{apalike}
\bibliography{References}
\end{document}